\documentclass{article} % For LaTeX2e
\usepackage{iclr2027_conference,times}
\iclrfinalcopy

\usepackage{amsmath,amsfonts,bm}

\def\eqref#1{equation~\ref{#1}}
\def\1{\bm{1}}

\DeclareMathAlphabet{\mathsfit}{\encodingdefault}{\sfdefault}{m}{sl}
\SetMathAlphabet{\mathsfit}{bold}{\encodingdefault}{\sfdefault}{bx}{n}

\usepackage[hidelinks]{hyperref}
\usepackage{url}
\usepackage[bottom]{footmisc}
\usepackage{graphicx}
\usepackage{booktabs}
\usepackage{tabularx}
\usepackage{multirow}
\usepackage{amssymb}
\usepackage{soul}
\usepackage{amsmath}
\usepackage{subcaption}
\usepackage{bbm}
\usepackage[x11names]{xcolor}
\usepackage{enumitem}

\definecolor{thoughtverylight}{HTML}{a8a495}
\definecolor{thoughtlight}{HTML}{89a0b0}
\definecolor{thoughtdark}{HTML}{607c8e}
\definecolor{thoughtedge}{HTML}{000000}
\definecolor{actionlight}{HTML}{c8aca9}
\definecolor{actiondark}{HTML}{c292a1}
\definecolor{actionedge}{HTML}{000000}
\definecolor{errThought}{HTML}{1F1F8F}
\definecolor{errAction}{HTML}{B8430B}
\definecolor{cClean}{HTML}{333333}
\definecolor{cSwap}{HTML}{0072B2}
\definecolor{cDrop}{HTML}{D55E00}
\definecolor{cPara}{HTML}{009E73}
\usepackage{listings}
\lstdefinestyle{promptbox}{
  basicstyle=\ttfamily\scriptsize, frame=single, framesep=4pt,
  breaklines=true, columns=fullflexible, keepspaces=true,
  showstringspaces=false, literate={`}{{\textasciigrave}}1
}

\usepackage{pgfplots}
\pgfplotsset{compat=1.18}
\usepgfplotslibrary{groupplots}
\usetikzlibrary{calc}
\usepackage{tikz}
\usepackage[most]{tcolorbox}
\usepackage{xcolor}
\usepackage{pifont}
\usetikzlibrary{calc}
\definecolor{thoughtred}{RGB}{170,55,55}
\definecolor{correctgreen}{RGB}{45,120,70}
\definecolor{obsgray}{RGB}{247,247,247}
\definecolor{wrongbg}{RGB}{253,245,245}
\definecolor{correctbg}{RGB}{245,250,246}
\definecolor{errTrain}{HTML}{1F4E9A}
\definecolor{errInfer}{HTML}{B8430B}
\newcommand{\std}[1]{$_{(#1)}$}
\newcommand{\cmark}{\textcolor{correctgreen}{\ding{51}}}
\newcommand{\xmark}{\textcolor{thoughtred}{\ding{55}}}
\title{Agents, Think Only When it Helps}

\usepackage{authblk}

\author{\textbf{Min-Seon Kim}$^{1,2}$ \quad \textbf{Seungdong Yoa}$^2$ \quad \textbf{Sanghyu Yoon}$^2$ \quad \textbf{Ye Seul Sim}$^2$ \\ \vspace{-0.1in}
       \textbf{Suhee Yoon}$^2$ \quad \textbf{Dongmin Kim}$^2$ \quad \textbf{Jung-Eun Kim}$^1$\thanks{Corresponding author.}
       \\ \vspace{0.1in}
       $^1$Computer Science, North Carolina State University\\
       \vspace{0.05in}
       $^2$LG AI Research\\
       }

\begin{document}

\maketitle

\begin{abstract}
LLM agent distillation transfers agentic capabilities to student models by training on complete teacher trajectories that typically interleave thoughts and actions. We question how critical thought supervision is for agents' ultimate action performance. To answer this, we disentangle their impacts by independently perturbing teacher thoughts and actions during distillation. On widely adopted agent benchmarks across model scales, altering or discarding thoughts has little effect on student performance. We find, however, that thought supervision becomes important when information about the thought is not disclosed in the corresponding action during training. By identifying such discrepant cases, our findings challenge the common practice of distilling entire teacher trajectories for LLM agents, and offer a new insight that thought supervision can be a discernible choice for efficiency depending on whether the information of thought is disclosed in the action.
\end{abstract}

\section{Introduction}

\begin{figure*}[b]
\vspace{-0.2cm}
\centering
\small

\begin{tcolorbox}[
    width=0.96\textwidth,
    colback=white,
    colframe=black!25,
    boxrule=0.6pt,
    arc=2pt,
    left=6pt,right=6pt,top=5pt,bottom=5pt
]
\textbf{Question.}
Marcus Wayne Garland played for a Baltimore-based team that was one of
the American League's eight charter teams.
In what year was this franchise founded?
\end{tcolorbox}

\begin{tcolorbox}[
    width=0.96\textwidth,
    colback=obsgray,
    colframe=black!15,
    boxrule=0.6pt,
    arc=2pt,
    left=6pt,right=6pt,top=5pt,bottom=5pt
]
\textbf{Observation}

\vspace{2pt}

\textbf{Document 0:}
``Baltimore Orioles ... As one of the American League's eight charter
teams \textbf{in 1901}, this particular franchise spent its first year
... in Milwaukee ...''

\vspace{2pt}

\textbf{Document 2:}
``... another AL charter member franchise also named the
`Baltimore Orioles,' which moved north \textbf{in 1903} to eventually
become the New York Yankees.''
\end{tcolorbox}

\begin{minipage}[t]{0.46\textwidth}
\begin{tcolorbox}[
    colback=wrongbg,
    colframe=thoughtred!60,
    boxrule=0.8pt,
    arc=2pt,
    height=1.9cm,
    valign=top,
    left=6pt,right=6pt,top=5pt,bottom=5pt
]
\centering
\textbf{\textcolor{thoughtred}{Student trained on Thoughts + Actions}}

\raggedright
\textbf{Thought:}
``The search results indicate that the Baltimore Orioles were
formed in \textbf{1903}.''

\vfill

\textbf{Final Answer:}
\textcolor{thoughtred}{\textbf{1903}} \hfill \xmark ~(Wrong)
\end{tcolorbox}
\end{minipage}
\hspace{0.03\textwidth}
\begin{minipage}[t]{0.46\textwidth}
\begin{tcolorbox}[
    colback=correctbg,
    colframe=correctgreen!60,
    boxrule=0.8pt,
    arc=2pt,
    height=1.9cm,
    valign=top,
    left=6pt,right=6pt,top=5pt,bottom=5pt
]
\centering
\textbf{\textcolor{correctgreen}{Student trained on Actions only}}

\vspace{5pt}

\raggedright
\textbf{Thought:}
\emph{(no thought generated)}

\vfill

\textbf{Final Answer:}
\textcolor{correctgreen}{\textbf{1901}} \hfill \cmark ~(Correct)
\end{tcolorbox}
\end{minipage}
\vspace{-0.1cm}
\caption{Two students distilled from the same teacher trajectories on HotpotQA~\citep{yang2018hotpotqa}, differing only in whether teacher thoughts are included in the supervision during training.}
\label{fig:motivating-example}

\end{figure*}

LLM agent distillation has emerged as a promising approach for transferring the capabilities of powerful teacher agents to student models through teacher-generated trajectories~\citep{hashemzadeh2024subgoal, qiu2025agentdistill, liu2026structured, kang2026distilling, lyu2025correction}. Existing agent distillation methods typically build on the ReAct framework~\citep{yao2023react}, in which trajectories interleave a natural-language \emph{thought}, an environment-facing \emph{action}, and the resulting \emph{observation}. Standard practice trains student models on complete teacher trajectories, jointly supervising both thoughts and actions.

On widely adopted agent benchmarks, thoughts dominantly occupy the trajectory budget. Nevertheless, we often observe that such a lengthy thought process does not always lead to correct reasoning and a correct answer, as shown in the example in Figure~\ref{fig:motivating-example}: a student trained on the complete trajectory follows an incorrect reasoning path and produces an incorrect answer, whereas a student trained without thoughts ends up with the correct one.

These observations make us question how critical thought supervision is for LLM agents' ultimate action performance. 
We examine this question by intervening in the thought and action channels, observing that corrupting teacher thoughts preserves near-clean student performance, whereas corrupting teacher actions causes substantial degradation. This asymmetry persists across student scales, teacher models, student model families, corruption granularities, and distillation objectives. On these benchmarks, we attribute this to the \textit{information of the thought being already disclosed in the action during training}, which can make thought supervision not essential or detrimental.

In contrast, we find that on quantitative reasoning benchmarks, thought information is not (or only a little) disclosed to the action during training, and thought supervision matters. In these settings, discarding teacher thoughts consistently degrades student performance, indicating that thought supervision leads to better reasoning and a greater likelihood of correct answers. To test whether this dependence arises from the disclosure of thought information conveyed by the actions, we increase the thought information disclosed to the action step without altering the underlying computation, and find that student models become less reliant on thought supervision.

Our central finding is that the importance of thought supervision depends on how actions are represented: specifically, on whether the information contained in the thought is already exposed in the action. When the thought information is already exposed in the action, the thought can be omitted with little or no loss in downstream performance---at times even improving it. When such information is not exposed, thought supervision yields more correct reasoning and answers. This challenges the default practice of indiscriminately distilling complete teacher trajectories, and instead motivates a more efficient, discerning use of thought supervision that preserves decision-relevant information while avoiding supervision on redundant thought tokens.

Our contributions are as follows:
\begin{enumerate}[]
    \item By disentangling thought and action supervision in agent distillation, we show that on widely adopted agent benchmarks, even corrupting or removing thoughts still largely preserves student performance.
    \item We find that the importance of thought supervision depends on how much the thought information is disclosed in the action.
    \item We show that, when the thought's information is already disclosed in the action, both training and inference cost can be saved by omitting thought supervision with little performance loss.
\end{enumerate}

\section{Related Work}
\label{sec:related}

\paragraph{Distillation for LLM Agents.}
Transferring agentic competence to a student model from a teacher model has converged on a common recipe: roll out a tool-using teacher in a ReAct-style loop~\citep{yao2023react} and fine-tune the student on the resulting thought--action--observation trajectories~\citep{chen2023fireact,kang2026distilling}. AgentTuning~\citep{zeng2024agenttuning} fine-tunes open LLMs on multi-task teacher trajectories in which every action is paired with a thought. Agent-FLAN~\citep{chen2024agent} attributes the inadequate agent learning of such corpora to the entanglement of format following and reasoning, and recasts them as multi-turn natural conversation. \citet{tang2026smartad} select teacher rollouts that match student capacity. 
A few methods depart from trajectory imitation, transferring reusable task modules without training on traces~\citep{qiu2025agentdistill} or replacing full teacher rollouts with student-generated trajectories that receive sparse corrections at their earliest errors~\citep{lyu2025correction}. 
Several studies rethink how teacher reasoning is used in agent distillation. \citet{hashemzadeh2024subgoal} replace free-form teacher reasoning with sub-goals over teacher trajectories and train a hierarchical agent whose planning and execution modules are supervised separately. \citet{penaloza2026privileged} study distillation when teacher reasoning is unavailable, using expert action trajectories as reference information for self-distillation. \citet{liu2026structured} ablate reasoning and action supervision, and \citet{liu2026cot} study whether chain-of-thought training improves direct action prediction or actions produced after reasoning. 
Unlike those studies, our work finds that the value of thought supervision varies with the extent to which thought information is exposed in the demonstrated action.

\paragraph{Reasoning Trace Supervision.}
Prior work has explored teacher-generated reasoning traces as an additional source of supervision for smaller models. Chain-of-thought distillation has shown that training students on intermediate rationales, rather than final answers alone, can improve reasoning performance and data efficiency~\citep{magister2023teaching, hsieh2023distilling, li2023symbolic}. However, subsequent work has questioned whether these gains require the student to faithfully reproduce or rely on the demonstrated reasoning process. For example, \citet{wadhwa2024investigating} find that rationale supervision can remain beneficial even when rationales are placed after the target labels, suggesting that its value does not necessarily arise from enabling step-by-step reasoning at inference time. \citet{chen2025unveiling} further show that the effectiveness of reasoning supervision varies with rationale granularity and student capacity. 
These findings suggest that reasoning traces can provide useful supervision, but leave open when and why they are critical. This question takes on a distinct form in agent distillation, where each teacher thought is paired with an environment-facing action that may already expose part of the information contained in the reasoning trace. We investigate how the information exposed in demonstrated actions relates to the value of thought supervision.

\section{Intervening on Thought and Action Supervision}
\label{sec:method}

\subsection{Agent Trajectories and Distillation}
We build on top of the ReAct paradigm~\citep{yao2023react} that existing agent distillation studies have widely adopted~\citep{hashemzadeh2024subgoal, qiu2025agentdistill, liu2026structured, kang2026distilling, tang2026smartad}. In this framework, a teacher trajectory \(\tau\) is an interleaved sequence of observations, thoughts, and actions,
\begin{equation}
\tau = (o_0, t_1, a_1, o_1, t_2, a_2, o_2, \ldots, t_T, a_T)
\label{eq:trajectory}
\end{equation}
where $o_i$ denotes the environment observation at step $i$, $t_i$ the natural language thought preceding the $i$-th action, $a_i$ the action executed in the environment, and $T$ the total number of decision steps. A trajectory is generated by a teacher policy $\pi_{\text{teacher}}$ and retained for distillation only if it reaches a successful terminal state, that is, $\text{Success}(\tau) = 1$ under the benchmark-specific success criterion. 
Then, we train the student parameters $\theta$ via supervised fine-tuning (SFT) using a standard autoregressive next-token prediction objective, with observation tokens \(o_i\) masked out from the loss.
\begin{equation}
\mathcal{L}_{\mathrm{SFT}}(\theta) = -\mathbb{E}_{\tau \sim \mathcal{D}}
\left[ 
\sum_{j \in \mathcal{S}(\tau)} \log \pi_\theta(y_j \mid y_{<j})
\right]
\label{eq:sft-loss}
\end{equation}
where $y_j$ denotes the $j$-th token in $\tau$, $\mathcal{S}(\tau)$ is the set of token positions corresponding to thoughts and actions, and $\mathcal{D}$ is the set of successful teacher trajectories. 
We additionally evaluate a KL-based distillation objective as an alternative training objective to test whether our findings are robust to the choice of distillation objective in Section~\ref{sec:exp-obj}.

\subsection{Corruption Operators}
\label{sec:intervention}

We perturb teacher thoughts while keeping the corresponding observations and actions fixed. Given a clean trajectory $\tau$, we define a corruption operator $C_{f,p} : \tau \mapsto \tau'_{f,p}$, where $f$ specifies the intervention and $p \in [0,1]$ controls the corruption ratio. For each thought $t_i$, we independently sample $z_i \sim \mathrm{Bernoulli}(p)$ and define
\begin{equation}
t_i' =
\begin{cases}
f(t_i), & \text{when } z_i = 1 \\
t_i, & \text{when } z_i = 0
\end{cases}
\end{equation}
We consider three interventions:
\begin{equation}
\begin{aligned}
\texttt{Drop: } & f_{\text{drop}}(t_i) = \varnothing  \\
\texttt{Replace: } & f_{\text{repl}}(t_i) = t^*, \quad t^* \sim \mathcal{D}^{t}_{\text{unrel}} \\
\texttt{Paraphrase: } & f_{\text{para}}(t_i) = \pi_{\text{paraphrase}}(t_i)
\end{aligned}
\label{eq:corruption-ops}
\end{equation}

\texttt{Drop} removes thought supervision from the training trajectory, \texttt{Replace} substitutes a thought with one from an unrelated task instance, and \texttt{Paraphrase} preserves its semantics while altering surface form. These interventions allow us to assess how omitting thought spans, replacing their content, and changing their wording affect the distilled student's performance. Here, $\mathcal{D}^{t}_{\text{unrel}}$ denotes the distribution of thoughts drawn from other task instances within the same benchmark, and $\pi_{\text{paraphrase}}$ denotes the paraphrasing model. We instantiate \(\pi_{\text{paraphrase}}\) with Qwen2.5-32B-Instruct, the same model used as our default teacher. The resulting corrupted trajectory is
\begin{equation}
\tau'_{f,p} = (o_0, t_1', a_1, o_1, \ldots, t_T', a_T)
\end{equation}

Replaced and paraphrased thought tokens remain supervised targets rather than being masked out, so the student is trained to generate them. At evaluation time, all students receive the ReAct-style prompts (Figures~\ref{fig:inference-prompt}--\ref{fig:inference-prompt-webshop}). Students trained with all thoughts dropped ($p{=}1$) nonetheless generate actions directly, as they have never observed thoughts during training.

Likewise, we intervene on the action channel while keeping all thoughts \(t_i\) and observations \(o_i\) fixed. We apply the same corruption procedure and corruption ratio \(p\) defined above to actions:
\begin{equation}
\begin{aligned}
\makebox[6.9em][r]{\texttt{Drop:}} & \quad
f_{\text{drop}}(a_i) = \varnothing \\
\makebox[6.9em][r]{\texttt{Replace:}} & \quad
f_{\text{repl}}(a_i) = a^*, \quad
a^* \sim \mathcal{D}^{a}_{\text{unrel}}
\end{aligned}
\label{eq:action-corruption}
\end{equation}
Here, $\mathcal{D}^{a}_{\text{unrel}}$ denotes the distribution of actions drawn from other task instances within the same benchmark. We do not apply paraphrasing to actions, as changes in surface form can alter their executable semantics.

\subsection{Experimental Setup}
Unless otherwise specified, teacher trajectories are generated by Qwen2.5-32B-Instruct~\citep{yang2024qwen25}, and students are initialized from Qwen2.5-3B-Instruct and trained with the SFT objective as in Eq.~\ref{eq:sft-loss}. To study robustness to student capacity, we evaluate Qwen2.5-Instruct students at four scales, 0.5\text{B}, 3\text{B}, 7\text{B}, and 14\text{B}. For each benchmark, all intervention conditions are constructed from the same underlying teacher trajectories and use identical training settings, differing only in the applied intervention. All students are evaluated under identical decoding settings, and results are averaged over three independent training and evaluation runs.

\paragraph{Evaluation Metrics.}
We evaluate each student model using task accuracy and its retention relative to the corresponding clean-trajectory baseline. Task accuracy is defined as
\begin{equation}
\text{Accuracy}(\pi_\theta, \mathcal{B})
=
100 \times \frac{1}{|\mathcal{B}|}
\sum_{x \in \mathcal{B}}
\mathbbm{1}\left[
\text{Success}\left(\tau_\theta(x)\right)=1
\right]
\label{eq:accuracy}
\end{equation}
where $\mathcal{B}$ denotes the benchmark evaluation set, $\tau_\theta(x)$ is the trajectory produced by rolling out the student policy $\pi_\theta$ on task instance $x$, and $\text{Success}(\cdot)$ denotes the benchmark-specific success criterion. For HotpotQA~\citep{yang2018hotpotqa}, 2WikiMultiHopQA~\citep{ho2020constructing}, Bamboogle~\citep{press2023measuring}, and MuSiQue~\citep{trivedi2022musique}, we use GPT-4o-mini~\citep{openai2024gpt4omini} as an LLM judge to assess semantic equivalence between the predicted and reference answers. Full evaluation details and the judge prompt are provided in Appendix~\ref{app:eval-detail}. Besides, to quantify performance degradation under corruption, we report \emph{retention}, defined as the accuracy of a student trained under intervention $(f,p)$, $\pi_{\theta}^{(f,p)}$, relative to that of the corresponding student trained on clean trajectories, $\pi_{\theta}$:
\begin{equation}
\text{Retention}(f,p,\mathcal{B})
=
100 \times
\frac{
\text{Accuracy}(\pi_{\theta}^{(f,p)}, \mathcal{B})
}{
\text{Accuracy}(\pi_{\theta}, \mathcal{B})
}
\label{eq:retention}
\end{equation}

\paragraph{Implementation Details.}
We fine-tune all student models for two epochs using LoRA with rank $64$, scaling factor $128$, dropout $0.05$, and all linear layers as target modules. We use AdamW with a learning rate of $1\times10^{-4}$, weight decay of $0.01$, and an effective batch size of $8$ trajectories. All experiments are conducted on a single NVIDIA H100 NVL GPU (96GB). Further details on training, inference, and evaluation are provided in Appendix~\ref{app:eval-detail}.

\section{Thought Supervision Is Often Not Critical on Common Agent Distillations}
\label{sec:experiments}

In this section, we examine how critical thought supervision is in agent distillation for downstream action performance across four benchmarks adopted from the original ReAct evaluation~\citep{yao2023react}: HotpotQA~\citep{yang2018hotpotqa} for multi-hop question answering, FEVER~\citep{thorne2018fever} for fact verification, WebShop~\citep{yao2022webshop} for web navigation, and ALFWorld~\citep{shridhar2021alfworld} for embodied decision-making. 
We directly compare the effect of corrupting the thought and action channels, and then test whether the resulting trends are consistent across model scales, model families, teacher models, corruption granularities, and distillation objectives. Detailed dataset statistics are provided in Table~\ref{tab:dataset_statistics} in Appendix~\ref{app:data}. Full numerical results for all experiments in Sections~\ref{sec:action-matters}--\ref{sec:ood} are tabulated in Appendix~\ref{app:exp}.

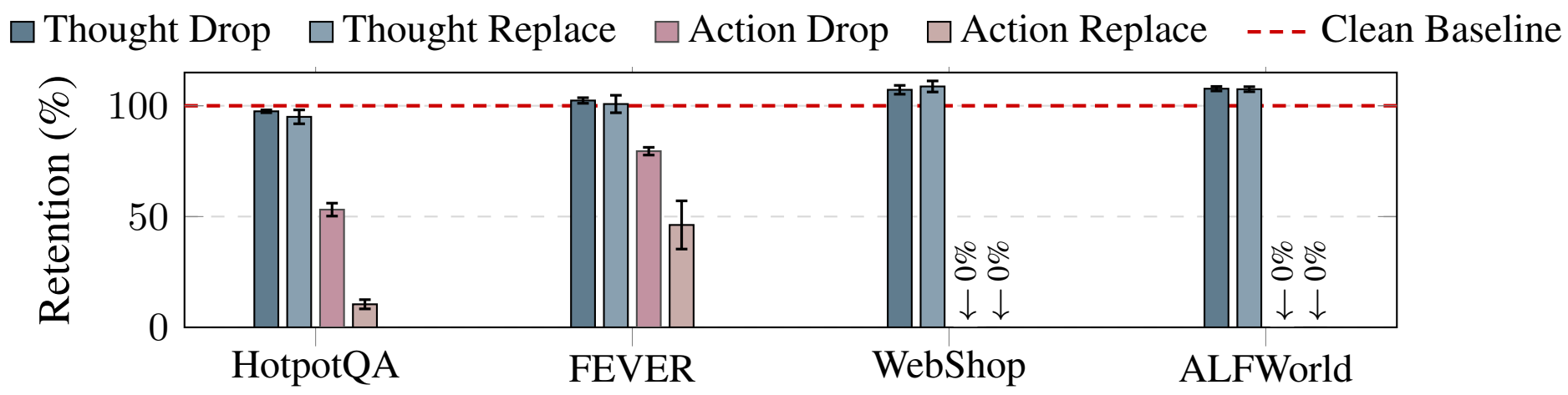
\begin{figure}[t]
\centering
\begin{tikzpicture}
\begin{axis}[
    ybar,
    width=0.8\linewidth,
    height=3.6cm,
    ymin=0, ymax=115,
    /pgf/bar width=5.4pt,
    ylabel={Retention (\%)},
    symbolic x coords={HotpotQA,FEVER,WebShop,ALFWorld},
    xtick=data,
    xticklabels={HotpotQA,FEVER,WebShop,ALFWorld},
    x tick label style={rotate=0, anchor=center, yshift=-5pt, font=\small},
    enlarge x limits=0.138,
    ymajorgrids=true,
    grid style={dashed,gray!30},
    tick label style={font=\small},
    extra y ticks={100},
    extra y tick labels={},
    extra y tick style={grid=major, major grid style={thick,red!80!black,densely dashed}},
    legend image code/.code={%
        \draw[#1, fill] (0cm,-0.08cm) rectangle (0.18cm,0.08cm);},
    legend style={
        at={(0.5,1.02)},
        anchor=south,
        draw=none,
        legend columns=-1,
        font=\small,
        /tikz/every even column/.append style={column sep=0.23cm}},
    every axis plot/.append style={
        error bars/y dir=both,
        error bars/y explicit,
    },
    thought err/.style={
        error bars/error bar style={line width=0.6pt, black},
        error bars/error mark options={rotate=90, mark size=1.3pt,
                                       line width=0.6pt, black},
    },
    action err/.style={
        error bars/error bar style={line width=0.6pt, black},
        error bars/error mark options={rotate=90, mark size=1.3pt,
                                       line width=0.6pt, black},
    },
    zero label/.style={
        nodes near coords={%
            \pgfmathfloatifflags{\pgfplotspointmeta}{0}{$\leftarrow$\,0\%}{}},
        every node near coord/.append style={
            font=\scriptsize,
            rotate=90,
            anchor=west,
            inner sep=1pt,
            black,
        },
    },
]
\addplot[fill=thoughtdark, draw=thoughtedge, thought err]
    coordinates {
        (HotpotQA,97.50)  +- (0,0.63)
        (FEVER,102.37)    +- (0,1.26)
        (WebShop,107.21)  +- (0,1.99)
        (ALFWorld,107.70) +- (0,1.02)
    };
\addlegendentry{Thought Drop}
\addplot[fill=thoughtlight, draw=thoughtedge, thought err]
    coordinates {
        (HotpotQA,95.00)  +- (0,3.13)
        (FEVER,100.79)    +- (0,3.94)
        (WebShop,108.71)  +- (0,2.49)
        (ALFWorld,107.47) +- (0,1.13)
    };
\addlegendentry{Thought Replace}
\addplot[fill=actiondark, draw=black!70, action err, zero label]
    coordinates {
        (HotpotQA,53.13)  +- (0,2.92)
        (FEVER,79.50)     +- (0,1.74)
        (WebShop,0.00)
        (ALFWorld,0.00)
    };
\addlegendentry{Action Drop}
\addplot[fill=actionlight, draw=actionedge, action err, zero label]
    coordinates {
        (HotpotQA,10.42)  +- (0,2.08)
        (FEVER,46.21)     +- (0,10.88)
        (WebShop,0.00)
        (ALFWorld,0.00)
    };
\addlegendentry{Action Replace}
\addlegendimage{legend image code/.code={%
    \draw[thick, red!80!black, densely dashed] (0cm,0cm) -- (0.5cm,0cm);}}
\addlegendentry{Clean Baseline}
\end{axis}
\end{tikzpicture}
\caption{Retention under thought and action corruption at $p{=}1.0$ (Qwen2.5-3B-Instruct student). Error bars denote standard deviation across three runs. Action drop and action replace yield 0\% retention on WebShop and ALFWorld.}
\label{fig:corruption_retention}
\end{figure}

\subsection{Actions Matter More Than Thoughts}
\label{sec:action-matters}
Figure~\ref{fig:corruption_retention} reports retention when the thought and action channels are independently corrupted with drop and replace at corruption ratio $p{=}1.0$. Action corruption causes substantial degradation: action drop retains 53.1\% of clean performance on HotpotQA and 79.5\% on FEVER, while action replacement retains only 10.4\% and 46.2\%, respectively. On FEVER, action replacement reduces task accuracy to 29.3\% (Table~\ref{tab:student-size}, Appendix~\ref{app:exp}), which is slightly below the chance level of 33.3\% for a three-choice question. On WebShop and ALFWorld, both action interventions yield 0.0\% retention. In contrast, thought drop and replacement preserve near-clean performance across all benchmarks, reaching 108.7\% on WebShop. 
We observe a clear distinction between the two channels: action corruption disrupts direct supervision of the observation-to-action mapping, whereas thought corruption leaves this mapping intact.
In particular, the robustness to thought replacement suggests that preserving the original thought content is not always necessary for transferring the demonstrated agent behavior. We also report various corruption ratios $p$ for both thought and action interventions in Figure~\ref{fig:corruption_sweep} (Appendix~\ref{app:exp}). 
This discrepancy suggests that action supervision in agent distillation is critical, while the thought channel is largely redundant.

\subsection{Robustness Across Student Scales}
\label{sec:student-scale}
\begin{figure}[t]
\centering
\pgfplotsset{
    every axis/.append style={
        ybar,
        bar width=4.5pt,
    }
}

% ============================================================
% (a) Across student model sizes
% ============================================================
\begin{subfigure}[t]{\linewidth}
\centering
\begin{tikzpicture}
\begin{groupplot}[
    group style={
        group size=4 by 1,
        horizontal sep=1.0cm,
    },
    ybar,
    width=4.0cm,
    height=3cm,
    /pgf/bar width=3.95pt,
    ymin=0,
    ymax=115,
    tick pos=left,
    symbolic x coords={0.5B,3B,7B,14B},
    xtick=data,
    x tick label style={font=\scriptsize},
    y tick label style={font=\scriptsize},
    ylabel style={font=\scriptsize},
    title style={font=\small},
    ymajorgrids=true,
    grid style={dashed,gray!25},
    extra y ticks={100},
    extra y tick labels={},
    extra y tick style={
        grid=major,
        major grid style={thick,red!80!black,densely dashed}
    },
    enlarge x limits=0.15,
    legend image code/.code={%
        \draw[#1, fill]
        (0cm,-0.08cm) rectangle (0.18cm,0.08cm);
    },
    every axis plot/.append style={
        error bars/y dir=both,
        error bars/y explicit,
        error bars/error bar style={line width=0.5pt, black},
        error bars/error mark options={rotate=90, mark size=1pt,
                                       line width=0.5pt, black},
    },
]

% ---------------- HotpotQA ----------------
\nextgroupplot[
    title={HotpotQA},
    ylabel={Retention (\%)},
    legend to name=corruptionlegend,
    legend columns=-1,
    legend style={
        draw=none,
        font=\small,
        /tikz/every even column/.append style={column sep=0.35cm}
    },
]
\addplot[fill=thoughtdark, draw=thoughtedge] coordinates {
    (0.5B,102.4) +- (0,1.80)
    (3B,97.5)    +- (0,0.63)
    (7B,106.3)   +- (0,1.18)
    (14B,103.9)  +- (0,1.60)
};
\addlegendentry{Thought Drop}
\addplot[fill=thoughtlight, draw=thoughtedge] coordinates {
    (0.5B,92.8)  +- (0,4.49)
    (3B,95.0)    +- (0,3.13)
    (7B,94.5)    +- (0,3.35)
    (14B,92.9)   +- (0,3.19)
};
\addlegendentry{Thought Replace}
\addplot[fill=thoughtverylight, draw=thoughtedge] coordinates {
    (0.5B,92.8)  +- (0,3.59)
    (3B,97.5)    +- (0,2.71)
    (7B,102.8)   +- (0,3.15)
    (14B,97.5)   +- (0,2.13)
};
\addlegendentry{Thought Paraphrase}
\addlegendimage{legend image code/.code={%
    \draw[thick, red!80!black, densely dashed] (0cm,0cm) -- (0.5cm,0cm);}}
\addlegendentry{Clean Baseline}

% ---------------- FEVER ----------------
\nextgroupplot[title={FEVER}]
\addplot[fill=thoughtdark, draw=thoughtedge] coordinates {
    (0.5B,95.3)  +- (0,1.56)
    (3B,102.4)   +- (0,1.26)
    (7B,102.1)   +- (0,1.36)
    (14B,94.0)   +- (0,1.58)
};
\addplot[fill=thoughtlight, draw=thoughtedge] coordinates {
    (0.5B,102.3) +- (0,2.34)
    (3B,100.8)   +- (0,3.94)
    (7B,103.0)   +- (0,2.26)
    (14B,95.4)   +- (0,1.72)
};
\addplot[fill=thoughtverylight, draw=thoughtedge] coordinates {
    (0.5B,98.8)  +- (0,0.98)
    (3B,97.8)    +- (0,1.89)
    (7B,98.8)    +- (0,1.66)
    (14B,94.6)   +- (0,1.29)
};

% ---------------- WebShop ----------------
\nextgroupplot[title={WebShop}]
\addplot[fill=thoughtdark, draw=thoughtedge] coordinates {
    (0.5B,94.4)  +- (0,0.93)
    (3B,107.2)   +- (0,1.99)
    (7B,101.9)   +- (0,1.69)
    (14B,105.8)  +- (0,0.97)
};
\addplot[fill=thoughtlight, draw=thoughtedge] coordinates {
    (0.5B,100.0) +- (0,2.09)
    (3B,108.7)   +- (0,2.49)
    (7B,101.4)   +- (0,3.14)
    (14B,100.0)  +- (0,3.38)
};
\addplot[fill=thoughtverylight, draw=thoughtedge] coordinates {
    (0.5B,92.1)  +- (0,1.86)
    (3B,94.0)    +- (0,2.24)
    (7B,106.3)   +- (0,3.38)
    (14B,101.4)  +- (0,2.66)
};

% ---------------- ALFWorld ----------------
\nextgroupplot[title={ALFWorld}]
\addplot[fill=thoughtdark, draw=thoughtedge] coordinates {
    (0.5B,109.9) +- (0,1.70)
    (3B,107.7)   +- (0,1.02)
    (7B,105.6)   +- (0,1.10)
    (14B,105.7)  +- (0,0.98)
};
\addplot[fill=thoughtlight, draw=thoughtedge] coordinates {
    (0.5B,107.8) +- (0,3.30)
    (3B,107.5)   +- (0,1.13)
    (7B,106.9)   +- (0,1.88)
    (14B,105.2)  +- (0,0.76)
};
\addplot[fill=thoughtverylight, draw=thoughtedge] coordinates {
    (0.5B,99.5)  +- (0,2.27)
    (3B,102.3)   +- (0,1.93)
    (7B,100.0)   +- (0,1.99)
    (14B,99.7)   +- (0,1.41)
};

\end{groupplot}

% Shared legend for the entire figure
\node[anchor=south] at
    ([yshift=17pt]$(group c1r1.north)!0.5!(group c4r1.north)$)
    {\pgfplotslegendfromname{corruptionlegend}};

\end{tikzpicture}
\caption{Across student model sizes.}
\label{fig:corruption_retention_size}
\end{subfigure}

\vspace{0.25cm}

% ============================================================
% (b) Across teacher models
% ============================================================
\begin{subfigure}[t]{\linewidth}
\vspace{-0.1cm}
\centering
\begin{tikzpicture}
\begin{groupplot}[
    group style={
        group size=4 by 1,
        horizontal sep=1.0cm,
    },
    ybar,
    width=4.0cm,
    height=3cm,
    /pgf/bar width=3.95pt,
    ymin=0,
    ymax=115,
    tick pos=left,
    symbolic x coords={qwen,gpt54,claude,gemini},
    xtick=data,
    xticklabels={
        {Qwen2.5-32B},
        {GPT-5.4},
        {Claude-Opus-5},
        {Gemini-3.1-Pro}
    },
    x tick label style={
        font=\tiny,
        rotate=25,
        anchor=east
    },
    y tick label style={font=\scriptsize},
    ylabel style={font=\scriptsize},
    title style={font=\small},
    ymajorgrids=true,
    grid style={dashed,gray!25},
    extra y ticks={100},
    extra y tick labels={},
    extra y tick style={
        grid=major,
        major grid style={thick,red!80!black,densely dashed}
    },
    enlarge x limits=0.15,
    every axis plot/.append style={
        error bars/y dir=both,
        error bars/y explicit,
        error bars/error bar style={line width=0.5pt, black},
        error bars/error mark options={rotate=90, mark size=1pt,
                                       line width=0.5pt, black},
    },
]

% ---------------- HotpotQA ----------------
\nextgroupplot[ylabel={Retention (\%)}]
\addplot[fill=thoughtdark, draw=thoughtedge] coordinates {
    (qwen,97.5)    +- (0,0.63)
    (gpt54,95.1)   +- (0,1.14)
    (claude,95.6)  +- (0,0.98)
    (gemini,97.2)  +- (0,1.40)
};
\addplot[fill=thoughtlight, draw=thoughtedge] coordinates {
    (qwen,95.0)    +- (0,3.13)
    (gpt54,92.8)   +- (0,2.85)
    (claude,93.7)  +- (0,3.41)
    (gemini,97.2)  +- (0,2.57)
};
\addplot[fill=thoughtverylight, draw=thoughtedge] coordinates {
    (qwen,97.5)    +- (0,2.71)
    (gpt54,99.6)   +- (0,2.66)
    (claude,100.5) +- (0,1.71)
    (gemini,109.3) +- (0,2.10)
};

% ---------------- FEVER ----------------
\nextgroupplot[]
\addplot[fill=thoughtdark, draw=thoughtedge] coordinates {
    (qwen,102.4)   +- (0,1.26)
    (gpt54,94.1)   +- (0,1.48)
    (claude,96.6)  +- (0,1.18)
    (gemini,97.3)  +- (0,1.53)
};
\addplot[fill=thoughtlight, draw=thoughtedge] coordinates {
    (qwen,100.8)   +- (0,3.94)
    (gpt54,93.4)   +- (0,3.61)
    (claude,95.3)  +- (0,2.87)
    (gemini,93.6)  +- (0,3.22)
};
\addplot[fill=thoughtverylight, draw=thoughtedge] coordinates {
    (qwen,97.8)    +- (0,1.74)
    (gpt54,96.4)   +- (0,2.30)
    (claude,99.3)  +- (0,1.69)
    (gemini,100.3) +- (0,1.86)
};
% ---------------- WebShop ----------------
\nextgroupplot[]
\addplot[fill=thoughtdark, draw=thoughtedge] coordinates {
    (qwen,107.2)   +- (0,1.99)
    (gpt54,100.8)  +- (0,1.88)
    (claude,107.4) +- (0,2.06)
    (gemini,105.4) +- (0,1.79)
};
\addplot[fill=thoughtlight, draw=thoughtedge] coordinates {
    (qwen,108.7)   +- (0,2.49)
    (gpt54,97.9)   +- (0,2.72)
    (claude,100.0) +- (0,2.47)
    (gemini,100.4) +- (0,2.91)
};
\addplot[fill=thoughtverylight, draw=thoughtedge] coordinates {
    (qwen,94.0)    +- (0,2.49)
    (gpt54,95.8)   +- (0,2.72)
    (claude,99.2)  +- (0,1.85)
    (gemini,103.6) +- (0,2.69)
};

% ---------------- ALFWorld ----------------
\nextgroupplot[]
\addplot[fill=thoughtdark, draw=thoughtedge] coordinates {
    (qwen,107.7)   +- (0,1.02)
    (gpt54,101.9)  +- (0,0.41)
    (claude,100.4) +- (0,0.40)
    (gemini,98.3)  +- (0,1.21)
};
\addplot[fill=thoughtlight, draw=thoughtedge] coordinates {
    (qwen,107.5)   +- (0,1.13)
    (gpt54,102.4)  +- (0,0.21)
    (claude,100.6) +- (0,0.20)
    (gemini,100.0) +- (0,0.40)
};
\addplot[fill=thoughtverylight, draw=thoughtedge] coordinates {
    (qwen,102.3)   +- (0,1.93)
    (gpt54,101.5)  +- (0,0.41)
    (claude,100.4) +- (0,0.40)
    (gemini,100.0) +- (0,0.40)
};
\end{groupplot}
\end{tikzpicture}
\caption{Across teacher models.}
\label{fig:corruption_retention_teacher}
\end{subfigure}

\vspace{0.25cm}

% ============================================================
% (c) Across student model families
% ============================================================
\begin{subfigure}[t]{\linewidth}
\vspace{-0.1cm}
\centering
\begin{tikzpicture}
\begin{groupplot}[
    group style={
        group size=4 by 1,
        horizontal sep=1.0cm,
    },
    ybar,
    width=4.0cm,
    height=3cm,
    /pgf/bar width=3.95pt,
    ymin=0,
    ymax=115,
    tick pos=left,
    symbolic x coords={llama,qwen,phi},
    xtick=data,
    xticklabels={
        {Llama-3.2},
        {Qwen2.5},
        {Phi-4-mini}
    },
    x tick label style={
        font=\tiny,
    },
    y tick label style={font=\scriptsize},
    ylabel style={font=\scriptsize},
    title style={font=\small},
    ymajorgrids=true,
    grid style={dashed,gray!25},
    extra y ticks={100},
    extra y tick labels={},
    extra y tick style={
        grid=major,
        major grid style={thick,red!80!black,densely dashed}
    },
    enlarge x limits=0.15,
    every axis plot/.append style={
        error bars/y dir=both,
        error bars/y explicit,
        error bars/error bar style={line width=0.5pt, black},
        error bars/error mark options={rotate=90, mark size=1pt,
                                       line width=0.5pt, black},
    },
]

% ---------------- HotpotQA ----------------
\nextgroupplot[ylabel={Retention (\%)}]
\addplot[fill=thoughtdark, draw=thoughtedge] coordinates {
    (llama,96.5)  +- (0,1.37)
    (qwen,97.5)   +- (0,0.63)
    (phi,93.4)    +- (0,2.20)
};
\addplot[fill=thoughtlight, draw=thoughtedge] coordinates {
    (llama,93.0)  +- (0,3.13)
    (qwen,95.0)   +- (0,3.13)
    (phi,93.4)    +- (0,4.41)
};
\addplot[fill=thoughtverylight, draw=thoughtedge] coordinates {
    (llama,100.8) +- (0,1.95)
    (qwen,97.5)   +- (0,1.46)
    (phi,104.0)   +- (0,2.86)
};

% ---------------- FEVER ----------------
\nextgroupplot[]
\addplot[fill=thoughtdark, draw=thoughtedge] coordinates {
    (llama,97.6)  +- (0,1.63)
    (qwen,102.4)  +- (0,1.26)
    (phi,100.9)   +- (0,2.31)
};
\addplot[fill=thoughtlight, draw=thoughtedge] coordinates {
    (llama,98.5)  +- (0,3.40)
    (qwen,100.8)  +- (0,3.94)
    (phi,96.0)    +- (0,4.78)
};
\addplot[fill=thoughtverylight, draw=thoughtedge] coordinates {
    (llama,95.0)  +- (0,1.78)
    (qwen,97.8)   +- (0,1.89)
    (phi,96.3)    +- (0,2.62)
};

% ---------------- WebShop ----------------
\nextgroupplot[]
\addplot[fill=thoughtdark, draw=thoughtedge] coordinates {
    (llama,105.4) +- (0,2.24)
    (qwen,107.2)  +- (0,1.99)
    (phi,100.7)   +- (0,3.08)
};
\addplot[fill=thoughtlight, draw=thoughtedge] coordinates {
    (llama,104.0) +- (0,2.91)
    (qwen,108.7)  +- (0,2.49)
    (phi,93.8)    +- (0,4.03)
};
\addplot[fill=thoughtverylight, draw=thoughtedge] coordinates {
    (llama,94.6)  +- (0,2.47)
    (qwen,94.0)   +- (0,2.24)
    (phi,94.3)    +- (0,3.32)
};

% ---------------- ALFWorld ----------------
\nextgroupplot[]
\addplot[fill=thoughtdark, draw=thoughtedge] coordinates {
    (llama,104.2) +- (0,1.17)
    (qwen,107.7)  +- (0,1.02)
    (phi,103.6)   +- (0,1.97)
};
\addplot[fill=thoughtlight, draw=thoughtedge] coordinates {
    (llama,103.8) +- (0,1.38)
    (qwen,107.5)  +- (0,1.13)
    (phi,104.4)   +- (0,2.40)
};
\addplot[fill=thoughtverylight, draw=thoughtedge] coordinates {
    (llama,98.4)  +- (0,1.17)
    (qwen,102.3)  +- (0,1.02)
    (phi,96.1)    +- (0,1.75)
};

\end{groupplot}
\end{tikzpicture}
\caption{Across student model families.}
\label{fig:corruption_retention_family}
\end{subfigure}

\vspace{-0.2cm}
\caption{
Retention under thought corruption across four benchmarks at $p=1.0$.
(a) Results across student model sizes.
(b) Results across teacher models.
(c) Results across student model families.
}
\label{fig:corruption_retention_robustness}
\end{figure}
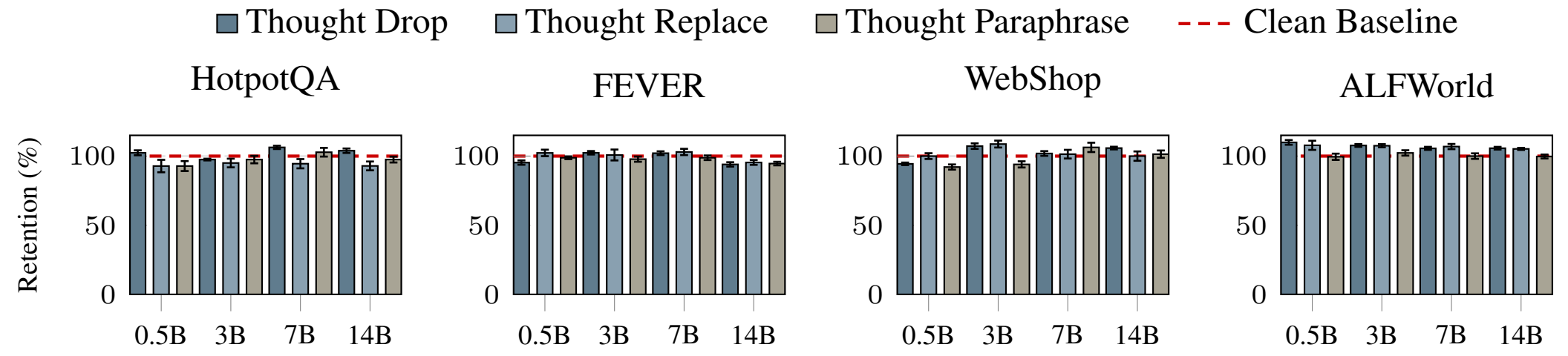
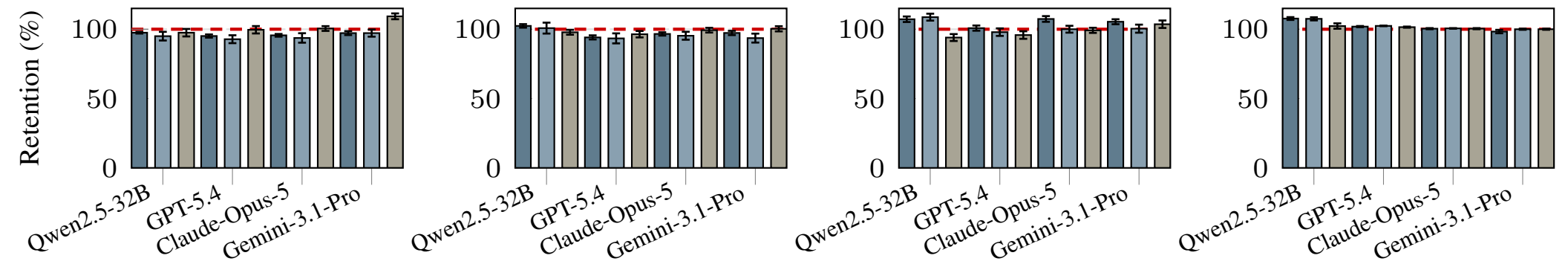
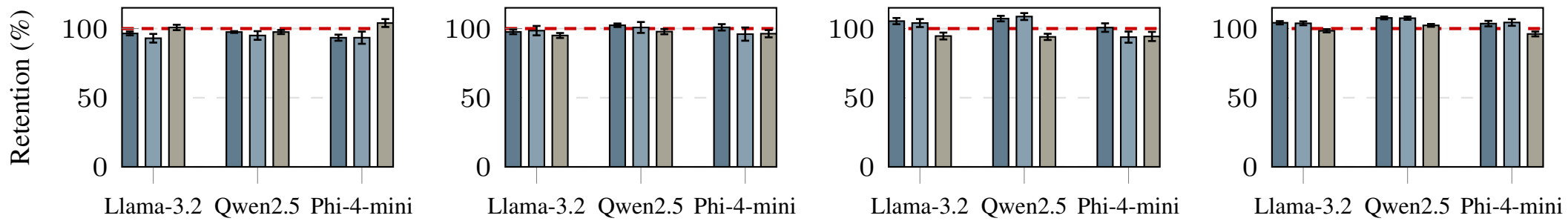

Figure~\ref{fig:corruption_retention_size} validates the insight across Qwen2.5-Instruct~\citep{yang2024qwen25} student scales of 0.5B, 3B, 7B, and 14B on all four benchmarks, reporting retention under thought replacement, removal, and paraphrase at $p{=}1.0$. Across all benchmarks, corruption operators, and student sizes, retention remains close to or above clean performance, ranging from 92.1\% to 109.9\%. This suggests that the limited utility of thought supervision is not a consequence of insufficient student model capacity.

\subsection{Robustness Across Teacher Models} 
\label{sec:teacher-model}
Figure~\ref{fig:corruption_retention_teacher} compares the default teacher, Qwen2.5-32B-Instruct~\citep{yang2024qwen25}, against three frontier teachers---GPT-5.4~\citep{openai2026gpt54}, Claude-Opus-5~\citep{anthropic2026opus5}, and Gemini-3.1-Pro~\citep{google2026gemini31}---reporting retention under thought interventions at $p{=}1.0$. Retention remains between 92.8\% and 109.3\% of clean performance regardless of the teacher used to generate the trajectories. 
These results show that robustness to thought interventions persists across diverse teacher models.

\subsection{Robustness Across Student Model Families} 
\label{sec:student-family}
Figure~\ref{fig:corruption_retention_family} distills the same Qwen2.5-32B-Instruct trajectories into three 3B-scale students---Llama-3.2-3B-Instruct~\citep{meta2024llama32}, Qwen2.5-3B-Instruct~\citep{yang2024qwen25}, and Phi-4-mini-instruct~\citep{abouelenin2025phi}. Retention ranges from 93.0\% to 108.7\% of clean performance, and thought drop slightly exceeds clean performance in the WebShop and ALFWorld settings. This suggests that the redundancy of thought supervision is not tied to a particular student model family.

\subsection{Robustness to Token-level Corruption}
\label{sec:token-level}
\begin{figure*}[t]
\centering
\begin{tikzpicture}
\begin{groupplot}[
    group style={group size=4 by 1, horizontal sep=0.95cm},
    width=0.29\textwidth, height=2.9cm,
    xtick={0,0.25,0.5,0.75,1.0},
    xticklabels={0,.25,.5,.75,1},
    xmin=-0.07, xmax=1.07,
    ymin=60, ymax=114,
    ytick={60,70,80,90,100,110},
    tick label style={font=\scriptsize},
    label style={font=\scriptsize},
    title style={font=\small},
    grid=major, grid style={black!12},
]

% ---------------- HotpotQA ----------------
\nextgroupplot[
    title=HotpotQA,
    ylabel={Retention (\%)},
    xlabel={corruption ratio $p$}
]
\addplot[thick,red!80!black,densely dashed, no marks, forget plot]
    coordinates {(-0.07,100) (1.07,100)};

\addplot[blue!55, mark=*] coordinates
    {(0,100) (0.25,100.4) (0.5,99.2) (0.75,95.0) (1.0,97.5)};

\addplot[orange!75!yellow, mark=square*] coordinates
    {(0,100) (0.25,99.6) (0.5,99.2) (0.75,93.1) (1.0,96.7)};

\addplot[green!55!black, mark=triangle*] coordinates
    {(0,100) (0.25,99.1) (0.5,98.2) (0.75,95.0) (1.0,93.2)};

% ---------------- FEVER ----------------
\nextgroupplot[
    title=FEVER,
    xlabel={corruption ratio $p$}
]
\addplot[thick,red!80!black,densely dashed, no marks, forget plot]
    coordinates {(-0.07,100) (1.07,100)};

\addplot[blue!55, mark=*] coordinates
    {(0,100) (0.25,103.8) (0.5,100.3) (0.75,100.3) (1.0,102.4)};

\addplot[orange!75!yellow, mark=square*] coordinates
    {(0,100) (0.25,98.4) (0.5,98.1) (0.75,98.6) (1.0,101.4)};

\addplot[green!55!black, mark=triangle*] coordinates
    {(0,100) (0.25,100.3) (0.5,98.4) (0.75,97.4) (1.0,96.1)};

% ---------------- WebShop ----------------
\nextgroupplot[
    title=WebShop,
    xlabel={corruption ratio $p$}
]
\addplot[thick,red!80!black,densely dashed, no marks, forget plot]
    coordinates {(-0.07,100) (1.07,100)};

\addplot[blue!55, mark=*] coordinates
    {(0,100) (0.25,103.5) (0.5,110.0) (0.75,99.5) (1.0,107.2)};

\addplot[orange!75!yellow, mark=square*] coordinates
    {(0,100) (0.25,103.0) (0.5,101.0) (0.75,107.5) (1.0,106.0)};

\addplot[green!55!black, mark=triangle*] coordinates
    {(0,100) (0.25,100) (0.5,97.9) (0.75,96.9) (1.0,96.9)};

% ---------------- ALFWorld ----------------
\nextgroupplot[
    title=ALFWorld,
    xlabel={corruption ratio $p$},
    legend to name=granlegend,
    legend columns=3,
    legend style={
        draw=none,
        fill=none,
        /tikz/every even column/.append style={column sep=0.15cm}
    }
]
\addplot[thick,red!80!black,densely dashed, no marks, forget plot]
    coordinates {(-0.07,100) (1.07,100)};

\addplot[blue!55, mark=*] coordinates
    {(0,100) (0.25,105.7) (0.5,108.4) (0.75,108.7) (1.0,107.7)};
\addlegendentry{Span (SFT)}

\addplot[orange!75!yellow, mark=square*] coordinates
    {(0,100) (0.25,104.5) (0.5,105.8) (0.75,105.8) (1.0,106.46)};
\addlegendentry{Token (SFT)}

\addplot[green!55!black, mark=triangle*] coordinates
    {(0,100) (0.25,97.8) (0.5,97.5) (0.75,100.0) (1.0,98.0)};
\addlegendentry{Span (KL)}
\end{groupplot}

\node[anchor=south] at
([yshift=14pt]$(group c1r1.north)!0.5!(group c4r1.north)$)
{\pgfplotslegendfromname{granlegend}};

\end{tikzpicture}
\caption{Retention under thought drop across corruption ratios $p$, comparing span-level and token-level corruption under SFT and span-level corruption under the KL objective (Qwen2.5-3B-Instruct student).}
\vspace{-0.2cm}
\label{fig:granularity}
\end{figure*}
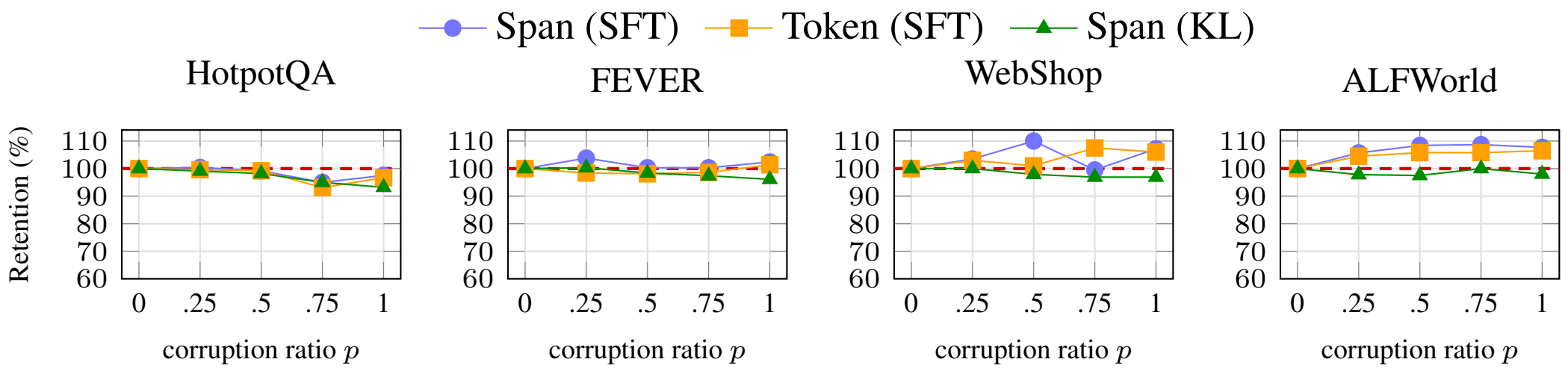

The preceding experiments corrupt each thought as a whole. We further test whether the observed robustness depends on corruption granularity by independently dropping each token within every thought with probability $p \in \{0.25, 0.5, 0.75, 1.0\}$. Figure~\ref{fig:granularity} shows that token- and span-level thought-drop produce similar retention patterns across all four benchmarks. Under token-level corruption, retention ranges from 93.1\% to 107.5\% of clean performance and averages 101.6\%, compared with 103.1\% under span-level corruption. These results show that robustness to thought corruption persists across corruption granularities. Additional results for token-level thought replacement and paraphrase are reported in Table~\ref{tab:span-vs-token} (Appendix~\ref{app:exp}).

\subsection{Robustness Across Distillation Objectives}
\label{sec:exp-obj}
To test whether our findings are specific to the SFT objective, we repeat the span-level thought-drop sweep under the KL-based distillation objective $\mathcal{L}_{\text{KL}}$ (Eq.~\ref{eq:kl-loss}, Appendix~\ref{app:eval-detail}). Figure~\ref{fig:granularity} shows that, at $p{=}1.0$, retention under KL-based distillation ranges from 93.2\% to 98.0\% across the four benchmarks, averaging 96.0\%, compared with an average of 103.7\% under SFT. These results suggest that the limited benefit of thought supervision is not specific to a particular distillation objective.

\subsection{Out-of-Distribution Robustness}
\label{sec:ood}

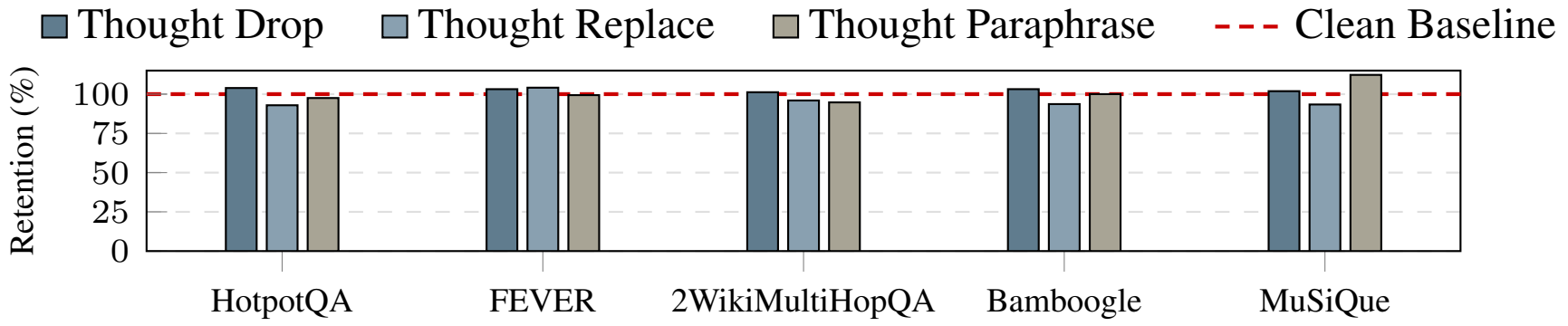
\begin{figure}[t]
\centering
\providecolor{thoughtdark}{RGB}{79,155,143}
\providecolor{thoughtlight}{RGB}{226,199,120}
\providecolor{thoughtrose}{RGB}{196,140,148}
\providecolor{thoughtedge}{RGB}{80,80,80}

\begin{tikzpicture}
\begin{axis}[
    ybar,
    width=0.8\linewidth,
    height=2.9cm,
    ymin=0, ymax=115,
    /pgf/bar width=6.4pt,
    ytick={0,25,50,75,100,125},
    tick pos=left,
    symbolic x coords={HotpotQA,FEVER,2WikiMultiHopQA,Bamboogle,MuSiQue},
    xtick=data,
    xticklabels={HotpotQA,FEVER,2WikiMultiHopQA,Bamboogle,MuSiQue},
    x tick label style={font=\scriptsize},
    y tick label style={font=\scriptsize},
    ylabel={Retention (\%)},
    ylabel style={font=\scriptsize},
    ymajorgrids=true,
    grid style={dashed,gray!25},
    extra y ticks={100},
    extra y tick labels={},
    extra y tick style={
        grid=major,
        major grid style={thick,red!80!black,densely dashed}
    },
    enlarge x limits=0.13,
    legend columns=-1,
    legend style={
        at={(0.5,1.03)},
        anchor=south,
        draw=none,
        font=\small,
        /tikz/every even column/.append style={column sep=0.35cm}
    },
    legend image code/.code={%
        \draw[#1,fill]
        (0cm,-0.08cm) rectangle (0.18cm,0.08cm);
    },
]

\addplot+[fill=thoughtdark, draw=thoughtedge, bar shift=-8.5pt]
coordinates {
    (HotpotQA,103.90)
    (FEVER,103.16)
    (2WikiMultiHopQA,101.21)
    (Bamboogle,103.17)
    (MuSiQue,101.89)
};
\addlegendentry{Thought Drop}

\addplot+[fill=thoughtlight, draw=thoughtedge, bar shift=0pt]
coordinates {
    (HotpotQA,92.91)
    (FEVER,104.11)
    (2WikiMultiHopQA,95.95)
    (Bamboogle,93.65)
    (MuSiQue,93.40)
};
\addlegendentry{Thought Replace}

\addplot+[fill=thoughtverylight, draw=thoughtedge, bar shift=8.5pt]
coordinates {
    (HotpotQA,97.52)
    (FEVER,99.37)
    (2WikiMultiHopQA,94.74)
    (Bamboogle,100.00)
    (MuSiQue,112.26)
};
\addlegendentry{Thought Paraphrase}

\addlegendimage{legend image code/.code={%
    \draw[thick,red!80!black,densely dashed]
    (0cm,0cm) -- (0.5cm,0cm);
}}
\addlegendentry{Clean Baseline}
\end{axis}
\end{tikzpicture}
\caption{Out-of-distribution robustness under thought interventions ($p{=}1.0$), with HotpotQA as in-distribution.}
\label{fig:cross-dataset-transfer}
\end{figure}

We evaluate the out-of-distribution robustness of thought interventions by distilling a Qwen2.5-14B-Instruct student on HotpotQA~\citep{yang2018hotpotqa} and evaluating it on FEVER~\citep{thorne2018fever}, 2WikiMultiHopQA~\citep{ho2020constructing}, Bamboogle~\citep{press2023measuring}, and MuSiQue~\citep{trivedi2022musique} without target-dataset fine-tuning. As shown in Figure~\ref{fig:cross-dataset-transfer}, thought interventions preserve performance across all out-of-distribution evaluation sets, retaining $93.4$--$112.3\%$ of clean performance. These results show that robustness to thought interventions largely extends to cross-dataset transfer.

\section{Thought Supervision Matters When Thought Is Not Disclosed in Action}
\label{sec:when-thought-matters}

\subsection{High-disclosure vs. Low-disclosure Settings}
\label{sec:high-vs-low}

In Section~\ref{sec:experiments}, we found that corrupting or even removing thoughts leaves performance largely unchanged, suggesting that thought supervision is often redundant for agents' action performance. 
Across all benchmarks above, we observed that information about the preceding thought is exposed in the action channel. In HotpotQA and FEVER, the action is a call to a web search tool, and the search query directly contains the entity or relation identified in the thought. In ALFWorld and WebShop, the action is a natural language command whose object names and attributes are constrained by the environment's valid action space, so the action accordingly contains this information. That is, the information introduced by the thought is disclosed verbatim or near-verbatim in the resulting action. We refer to such settings as \emph{high-disclosure}. Such disclosure makes thought supervision not very useful because much of the task-relevant information it provides is already conveyed by the action.

Disclosure is much lower when actions are the symbolic programs provided by FinQA~\citep{chen2021finqa}, MultiHiertt~\citep{zhao2022multihiertt}, MathQA~\citep{amini2019mathqa}, and GeoQA~\citep{chen2021geoqa}. These programs consist of short operations over anonymous intermediate variables; we convert them into a unified step-wise executable format while preserving their original computation and anonymous references. The resulting action specifies how to compute the answer, but not what each operand represents or why that computation is appropriate. That information appears only in the accompanying thought, generated by the Qwen2.5-32B-Instruct teacher, leaving room for thought supervision to provide additional benefits beyond the actions. We refer to these as \emph{low-disclosure} settings. 
Figure~\ref{fig:disclosure-examples} shows contrasting examples from the two different settings. In Figure~\ref{fig:disclosure-examples} (a), taken from FEVER, parts of information introduced in the thought are repeated almost verbatim in the action's search query, creating the disclosure described above. In Figure~\ref{fig:disclosure-examples} (b), taken from MathQA, the action performs the arithmetic operations based on the thought but does not reveal what each quantity means or why those quantities should be combined in that way.

\colorlet{qleak}{orange!45}
\colorlet{aleak}{blue!23}
\newcommand{\qhl}[1]{{\sethlcolor{qleak}\hl{#1}}}
\newcommand{\ahl}[1]{{\sethlcolor{aleak}\hl{#1}}}
\newcommand{\fieldrow}[2]{%
  \noindent\textbf{\footnotesize\textsf{#1}}\ #2\par\vspace{2pt}%
}
\begin{figure}[t]
\centering
\begin{tcolorbox}[
  enhanced, unbreakable, width=\linewidth,
  colback=blue!2, colframe=blue!55, boxrule=0.4pt, arc=1mm,
  fonttitle=\bfseries\footnotesize,
  title={(a) \textsf{THOUGHT} is disclosed in \textsf{ACTION} (FEVER)}, top=3mm, left=2mm, right=2mm, bottom=2mm]
\footnotesize
\fieldrow{QUESTION}{\,\,Claim: ``Winter's Tale is a 1983 novel.'' Verify using evidence from Wikipedia.}\vspace{0.05cm}
\fieldrow{THOUGHT}{\,\,I will use the tool \texttt{web\_search} to get the information about \ahl{Winter's Tale} and its \ahl{publication year}.}\vspace{0.05cm}
\fieldrow{ACTION}{\,\,\texttt{evidence = web\_search(query="\ahl{Winter's Tale} \ahl{publication year}")}\newline\texttt{print(evidence)}}
\end{tcolorbox}
\begin{tcolorbox}[
  enhanced, unbreakable, width=\linewidth,
  colback=orange!5, colframe=orange!75!yellow, boxrule=0.6pt, arc=1mm,
  fonttitle=\bfseries\footnotesize,
  title={(b) \textsf{THOUGHT} is \emph{not} disclosed in \textsf{ACTION} (MathQA)}, top=3mm, left=2mm, right=2mm, bottom=2mm]
\footnotesize
\fieldrow{QUESTION}{\,\,A shop owner sells 15 mtr of cloth and gains sp of 10 mtrs. Find the gain \%?}\vspace{0.05cm}
\fieldrow{THOUGHT}{\,\,To find the gain percentage, I need to calculate the profit made on the cost price, which is the difference between the selling price of 15 meters and the gain of 10 meters.}\vspace{0.05cm}
\fieldrow{ACTION}{\,\,\texttt{step0 = 15 - 10}\newline
\texttt{step1 = 10 / step0}\newline
\texttt{print(step1 * 100.0)}}
\end{tcolorbox}
\caption{Examples illustrating whether task-relevant information of the thought is disclosed in the action during training. (a) Parts of thought information are disclosed in the action; (b) Thought is not disclosed in the action.}
\label{fig:disclosure-examples}
\end{figure}

\subsection{Removing Thought in Low-disclosure Settings}
\label{sec:low-disclosure}

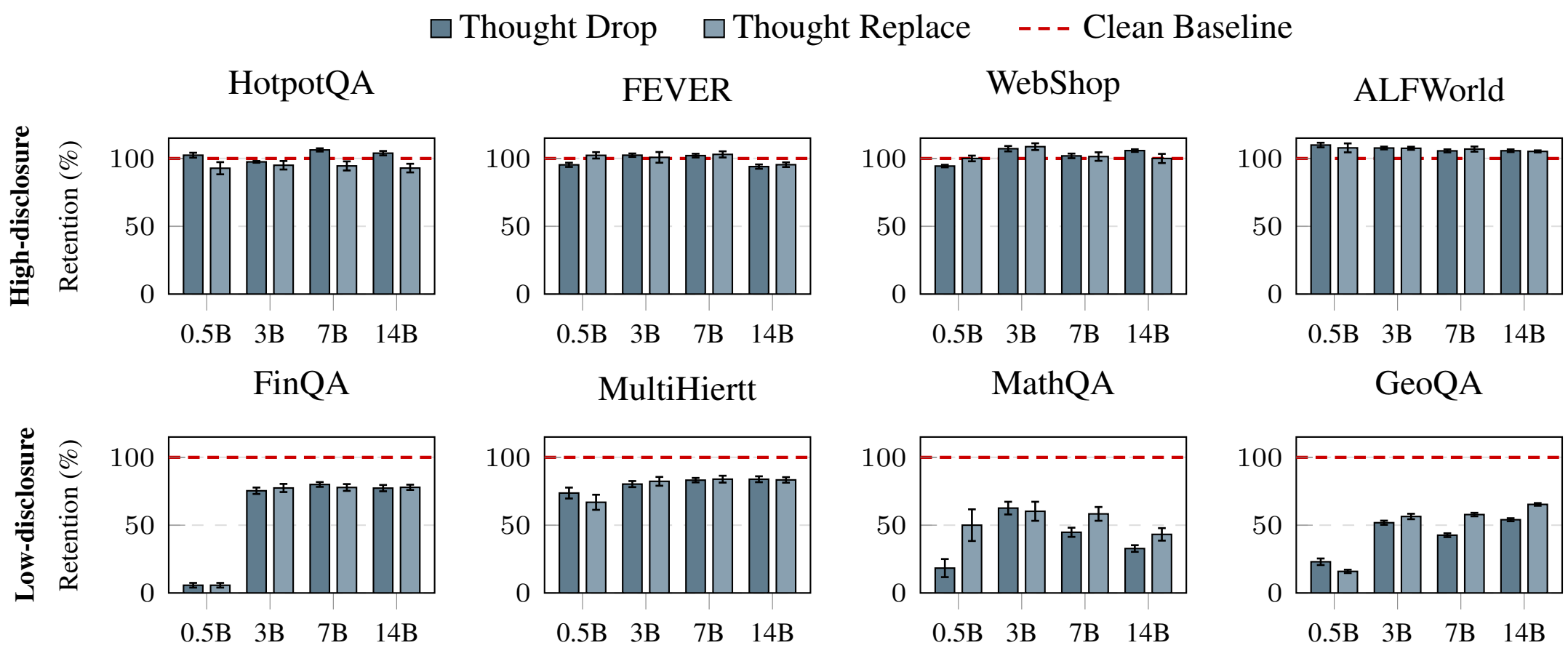
\begin{figure}[t]
\centering
\pgfplotsset{
    every axis/.append style={ybar},
    witherr/.style={
        error bars/y dir=both,
        error bars/y explicit,
        error bars/error bar style={line width=0.5pt, black},
        error bars/error mark options={rotate=90, mark size=1pt,
                                       line width=0.5pt, black},
    },
}
\begin{tikzpicture}
\begin{groupplot}[
    group style={
        group size=4 by 2,
        horizontal sep=1.0cm,
        vertical sep=1.3cm,
    },
    width=4cm,
    height=3cm,
    /pgf/bar width=5pt,
    ymin=0,
    ymax=115,
    tick pos=left,
    symbolic x coords={0.5B,3B,7B,14B},
    xtick=data,
    x tick label style={font=\scriptsize},
    y tick label style={font=\scriptsize},
    ylabel style={font=\scriptsize},
    title style={font=\small},
    ymajorgrids=true,
    grid style={dashed,gray!25},
    extra y ticks={100},
    extra y tick labels={},
    extra y tick style={grid=major, major grid style={thick,red!80!black,densely dashed}},
    enlarge x limits=0.20,
    legend image code/.code={%
        \draw[#1, fill]
        (0cm,-0.08cm) rectangle (0.18cm,0.08cm);
    },
]
\nextgroupplot[title={HotpotQA}, ylabel={Retention (\%)}]
\addplot+[fill=thoughtdark, draw=thoughtedge, witherr] coordinates {
    (0.5B,102.4) +- (0,1.80)
    (3B,97.5)    +- (0,0.63)
    (7B,106.3)   +- (0,1.18)
    (14B,103.9)  +- (0,1.60)
};
\addplot+[fill=thoughtlight, draw=thoughtedge, witherr] coordinates {
    (0.5B,92.8)  +- (0,4.49)
    (3B,95.0)    +- (0,3.13)
    (7B,94.5)    +- (0,3.35)
    (14B,92.9)   +- (0,3.19)
};

\nextgroupplot[title={FEVER}]
\addplot+[fill=thoughtdark, draw=thoughtedge, witherr] coordinates {
    (0.5B,95.3)  +- (0,1.56)
    (3B,102.4)   +- (0,1.26)
    (7B,102.1)   +- (0,1.36)
    (14B,94.0)   +- (0,1.58)
};
\addplot+[fill=thoughtlight, draw=thoughtedge, witherr] coordinates {
    (0.5B,102.3) +- (0,2.34)
    (3B,100.8)   +- (0,3.94)
    (7B,103.0)   +- (0,2.26)
    (14B,95.4)   +- (0,1.72)
};

\nextgroupplot[title={WebShop}]
\addplot+[fill=thoughtdark, draw=thoughtedge, witherr] coordinates {
    (0.5B,94.4)  +- (0,0.93)
    (3B,107.2)   +- (0,1.99)
    (7B,101.9)   +- (0,1.69)
    (14B,105.8)  +- (0,0.97)
};
\addplot+[fill=thoughtlight, draw=thoughtedge, witherr] coordinates {
    (0.5B,100.0) +- (0,2.09)
    (3B,108.7)   +- (0,2.49)
    (7B,101.4)   +- (0,3.14)
    (14B,100.0)  +- (0,3.38)
};

\nextgroupplot[title={ALFWorld}]
\addplot+[fill=thoughtdark, draw=thoughtedge, witherr] coordinates {
    (0.5B,109.9) +- (0,1.70)
    (3B,107.7)   +- (0,1.02)
    (7B,105.6)   +- (0,1.10)
    (14B,105.7)  +- (0,0.98)
};
\addplot+[fill=thoughtlight, draw=thoughtedge, witherr] coordinates {
    (0.5B,107.8) +- (0,3.30)
    (3B,107.5)   +- (0,1.13)
    (7B,106.9)   +- (0,1.88)
    (14B,105.2)  +- (0,0.76)
};

\nextgroupplot[
    title={FinQA},
    ylabel={Retention (\%)},
    legend to name=corruptionlegend,
    legend columns=-1,
    legend style={
        draw=none,
        font=\small,
        /tikz/every even column/.append style={column sep=0.35cm}
    },
]
\addplot+[fill=thoughtdark, draw=thoughtedge, witherr] coordinates {
    (0.5B,5.69)  +- (0,1.71)
    (3B,75.36)   +- (0,2.37)
    (7B,80.00)   +- (0,1.76)
    (14B,77.30)  +- (0,2.32)
};
\addlegendentry{Thought Drop}
\addplot+[fill=thoughtlight, draw=thoughtedge, witherr] coordinates {
    (0.5B,5.69) +- (0,1.71)
    (3B,77.41)   +- (0,3.00)
    (7B,77.79)   +- (0,2.50)
    (14B,77.9)  +- (0,1.94)
};
\addlegendentry{Thought Replace}
\addlegendimage{legend image code/.code={%
    \draw[thick, red!80!black, densely dashed] (0cm,0cm) -- (0.5cm,0cm);}}
\addlegendentry{Clean Baseline}

\nextgroupplot[title={MultiHiertt}]
\addplot+[fill=thoughtdark, draw=thoughtedge, witherr] coordinates {
    (0.5B,73.68) +- (0,4.02)
    (3B,80.28)   +- (0,2.24)
    (7B,83.18)   +- (0,1.63)
    (14B,83.86)  +- (0,2.17)
};
\addplot+[fill=thoughtlight, draw=thoughtedge, witherr] coordinates {
    (0.5B,66.87) +- (0,5.57)
    (3B,82.32)   +- (0,3.25)
    (7B,83.91)   +- (0,2.53)
    (14B,83.36)  +- (0,2.01)
};

\nextgroupplot[title={MathQA}]
\addplot+[fill=thoughtdark, draw=thoughtedge, witherr] coordinates {
    (0.5B,18.33) +- (0,6.67)
    (3B,62.57)   +- (0,4.68)
    (7B,44.75)   +- (0,3.39)
    (14B,32.77)  +- (0,2.43)
};
\addplot+[fill=thoughtlight, draw=thoughtedge, witherr] coordinates {
    (0.5B,50.00) +- (0,11.67)
    (3B,60.23)   +- (0,7.02)
    (7B,58.31)   +- (0,5.08)
    (14B,43.20)  +- (0,4.61)
};

\nextgroupplot[title={GeoQA}]
\addplot+[fill=thoughtdark, draw=thoughtedge, witherr] coordinates {
    (0.5B,23.0) +- (0,2.41)
    (3B,51.8)   +- (0,1.52)
    (7B,42.6)   +- (0,1.40)
    (14B,54.0)  +- (0,1.12)
};
\addplot+[fill=thoughtlight, draw=thoughtedge, witherr] coordinates {
    (0.5B,15.9) +- (0,1.22)
    (3B,56.4)   +- (0,2.00)
    (7B,57.8)   +- (0,1.27)
    (14B,65.3)  +- (0,0.99)
};

\end{groupplot}
\node[rotate=90, anchor=south, font=\scriptsize\bfseries]
    at ([xshift=-1.1cm]group c1r1.west) {High-disclosure};
\node[rotate=90, anchor=south, font=\scriptsize\bfseries]
    at ([xshift=-1.1cm]group c1r2.west) {Low-disclosure};

\node[anchor=south] at
    ([yshift=17pt]$(group c1r1.north)!0.5!(group c4r1.north)$)
    {\pgfplotslegendfromname{corruptionlegend}};

\end{tikzpicture}
\caption{Retention under thought drop and replace across student sizes for high- and low-disclosure settings ($p{=}1.0)$.}
\label{fig:thought-drop-contrast}
\end{figure}

To examine whether thought supervision becomes important when thoughts are not disclosed in the action, we ablate thoughts from the teacher trajectories during distillation. Figure~\ref{fig:thought-drop-contrast} shows a clear contrast between the two regimes: thought drop substantially impairs agent performance in low-disclosure settings, whereas performance remains largely preserved on high-disclosure settings across all model scales. For the 14B student in low-disclosure settings, retention drops to 77.3\% on FinQA, 83.9\% on MultiHiertt, 32.8\% on MathQA, and 54.0\% on GeoQA. In contrast to the high-disclosure settings, where thought removal largely preserves or even improves performance, low-disclosure settings consistently benefit from thought supervision in which task-relevant information of the thought is not exposed in the action. 
This pattern provides correlational evidence that thought supervision becomes more important as less thought information is disclosed in the action. We further control for inference-time thought generation and find that full-trajectory-trained students continue to outperform action-only students even when thoughts are suppressed at inference, showing that this benefit cannot be explained merely by differences in inference-time reasoning (Appendix~\ref{app:inference_control}). Full results are provided in Table~\ref{tab:finqa-thought-drop} (Appendix~\ref{app:exp}).

\subsection{Ablation: Disclosing Thought in Action}
\label{sec:disclose}

To test whether increasing the amount of thought information disclosed in the action affects the effect of thought supervision, we directly intervene in the action format of the low-disclosure settings by increasing the amount of thought information disclosed in the action while preserving the underlying computation. In low-disclosure settings, the actions contain anonymous intermediate references that specify how to compute the answer but reveal little about what each quantity represents. We therefore create a more disclosed variant by replacing these anonymous variables with semantic names that describe the quantities they hold. The numeric values, operators, execution order, and final outputs are kept unchanged; only the degree to which the action reveals the reasoning-relevant content is increased. 
Figure~\ref{fig:disclosure-intervention-main} shows an example from MathQA, before and after the disclosure intervention. In the original trace, the thought identifies \texttt{cost\_price} as the quantity that \texttt{15 - 10} denotes, but the action exposes none of these: \texttt{step0} is semantically empty, so the mapping from the question's entities to the computed quantities is recoverable only from the thought. After renaming, that mapping is carried by the identifiers themselves, and the action explicitly discloses content that previously existed only in the thought.

\colorlet{qleak}{gray!30}
\colorlet{aleak}{gray!30}
\newcommand{\whl}[1]{{\setlength{\fboxsep}{1pt}\colorbox{aleak}{\texttt{#1}}}}
\begin{figure}[t]
\centering
\begin{tcolorbox}[
  enhanced, unbreakable, width=\linewidth,
  colback=gray!5, colframe=gray!80!yellow, boxrule=0.4pt, arc=1mm,
  fonttitle=\bfseries\footnotesize,
  title={\textsf{After artificially disclosing}---\textsf{THOUGHT} is disclosed in \textsf{ACTION} (MathQA)},
  top=3mm, left=2mm, right=2mm, bottom=2mm]
\footnotesize
\fieldrow{QUESTION}{\,\,A shop owner sells 15 mtr of cloth and gains sp of 10 mtrs. Find the gain \%?}\vspace{0.05cm}
\fieldrow{THOUGHT}{\,\,To find the gain percentage, I need to calculate the profit made on the \ahl{cost price}, which is the \ahl{difference between the selling price of 15 meters} and the \ahl{gain of 10 meters}.}\vspace{0.05cm}
\fieldrow{ACTION}{\,\,\whl{cost\_price}\texttt{ = 15 - 10}\newline
\whl{gain\_over\_cost\_price}\texttt{ = 10 / }\whl{cost\_price}\newline
\texttt{print(}\whl{gain\_over\_cost\_price}\texttt{ * 100.0)}}
\end{tcolorbox}
\vspace{-0.1in}
\caption{Example of the disclosure intervention applied to the MathQA instance in  Figure~\ref{fig:disclosure-examples}(b). Anonymous variables in the action are renamed with semantic names, exposing information that was previously available only in the thought.}
\label{fig:disclosure-intervention-main}
\end{figure}

\begin{figure}[b]
\centering
\begin{minipage}[c]{0.45\textwidth}
\centering
\begin{tikzpicture}
\begin{axis}[
    ybar,
    width=\linewidth,
    height=3.5cm,
    bar width=8pt,
    enlarge x limits=0.15,
    symbolic x coords={FinQA, MultiHiertt, MathQA, GeoQA},
    xtick=data,
    ylabel={Retention (\%)},
    ylabel style={font=\scriptsize},
    tick label style={font=\scriptsize},
    ymin=40, ymax=110,
    ytick={40,60,80,100},
    ymajorgrids=true,
    grid style={dashed,gray!25},
    tick pos=left,
    legend image code/.code={%
        \draw[#1, fill] (0cm,-0.08cm) rectangle (0.18cm,0.08cm);},
    legend style={
        at={(0.5,1.02)}, anchor=south,
        draw=none, legend columns=3, font=\scriptsize,
        /tikz/every even column/.append style={column sep=0.15cm}},
]
    \addplot+[fill=gray!30, draw=black!45, bar shift=-4.5pt]
        coordinates {(FinQA,75.4) (MultiHiertt,80.3) (MathQA,62.6) (GeoQA,51.8)};
    \addlegendentry{Before disclosing}

    \addplot+[fill=blue!55!black, draw=blue!75!black, bar shift=4.5pt]
        coordinates {(FinQA,94.6) (MultiHiertt,88.1) (MathQA,91.3) (GeoQA,84.4)};
    \addlegendentry{After disclosing}

    \addlegendimage{legend image code/.code={%
        \draw[thick, red!80!black, densely dashed] (0cm,0cm) -- (0.4cm,0cm);}}
    \addlegendentry{Clean Baseline}
    \draw[thick, densely dashed, red!80!black]
        ({rel axis cs:0,0}|-{axis cs:FinQA,100}) -- ({rel axis cs:1,0}|-{axis cs:FinQA,100});
\end{axis}
\end{tikzpicture}
\captionof{figure}{Thought-drop retention before and after artificially disclosing thought information in low-disclosure settings (Qwen2.5-3B-Instruct student).}
\label{fig:disclosure_intervention}
\end{minipage}\hfill
\begin{minipage}[c]{0.52\textwidth}
\centering
\captionof{table}{Accuracy before and after artificially disclosing thought information in low-disclosure settings ($p{=}1.0$). (The same data points in Figure~\ref{fig:disclosure_intervention})}
\label{tab:action-disclosure-main}
\small
\resizebox{\linewidth}{!}{
\begin{tabular}{lcccc}
\toprule
& \multicolumn{2}{c}{Before disclosing} & \multicolumn{2}{c}{After disclosing} \\
\cmidrule(lr){2-3}\cmidrule(lr){4-5}
Benchmark & Clean & Drop & Clean & Drop \\
\midrule
FinQA       & 63.3\std{0.7} & 47.7\std{1.5} & 65.3\std{0.7} & 61.8\std{1.2} \\
MultiHiertt & 49.2\std{0.6} & 39.5\std{1.1} & 50.6\std{0.6} & 44.6\std{1.3} \\
MathQA      & 17.1\std{0.9} & 10.7\std{0.8} & 18.3\std{0.8} & 16.7\std{0.9} \\
GeoQA       & 38.8\std{0.6} & 20.1\std{0.5} & 41.0\std{0.8} & 34.6\std{0.8} \\
\bottomrule
\end{tabular}
}
\end{minipage}
\end{figure}

Figure~\ref{fig:disclosure_intervention} shows that exposing more thought information in the action reduces the student's dependence on the thought channel during distillation. Before the intervention, retention ranges from $51.8$--$80.3\%$, well below $97.5$--$107.7$\% observed on high-disclosure settings. After the intervention, retention consistently improves on all four benchmarks, rising to $84.4$--$94.6$\%. Importantly, this improvement is not an artifact of a reduced clean baseline: as shown in Table~\ref{tab:action-disclosure-main}, clean accuracy actually increases slightly across all four benchmarks, while thought-drop accuracy increases substantially more. 
Once task-relevant semantics are partially revealed in the action itself, removing the thought becomes substantially less harmful. Because the underlying computation remains unchanged, this improvement reflects a change in the information available through action supervision. 
Taken together with the earlier comparisons across settings, this supports our central claim: thought supervision becomes less useful as more thought information is disclosed in the action. 
To verify that this gain stems from semantic correspondence of the identifiers rather than from the mere presence of thought-derived words, we also evaluate names that are derived from the thoughts but do not describe the quantities they represent (Appendix~\ref{app:disclosure_control}).

\section{Bypassing Thought Supervision for Efficient Agent Distillation}
\label{sec:solution}

By recognizing and distinguishing between high-disclosure and low-disclosure settings, we find that thought supervision provides no clear benefit in high-disclosure settings and can often be omitted from them with little or no loss in performance. The distinction motivates an alternative approach to supervision for such settings: distilling action only in order to gain efficiency in both training and inference.

In the evaluated high-disclosure settings, omitting thoughts improves training and inference efficiency while largely preserving, and at times improving, task performance. Thought tokens account for $59.1$--$71.5\%$ of assistant-generated tokens across four benchmarks and four student scales (Figure~\ref{fig:token_reduction}). Removing thought tokens shortens the training sequences, which reduces the computation required per example during training. Students trained without thoughts generate actions directly in response to the same evaluation prompts, so the choice of training supervision also affects inference efficiency. Relative to full-trajectory distillation, action-only distillation reduces training wall-clock time by $5.3$--$10.2\%$ and inference wall-clock time by $11.1$--$41.7\%$ across the four benchmarks (Figure~\ref{fig:wallclock_reduction}). These savings show a practical benefit of omitting redundant thought supervision during both distillation and deployment, while largely maintaining model utility.

\begin{figure*}[h]
\centering
% ============================================================
% (a) Token reduction
% ============================================================
\begin{minipage}[t]{0.6\textwidth}
\centering
\begin{tikzpicture}[baseline=(tokax.south)]
\begin{axis}[
    name=tokax,
    ybar stacked,
    bar width=8pt,
    scale only axis,
    width=0.88\linewidth,
    height=3cm,
    ylabel={Assistant tokens / trajectory},
    ylabel style={font=\small},
    tick label style={font=\small},
    ymin=0, ymax=400,
    symbolic x coords={
        H05,H3,H7,H14,
        F05,F3,F7,F14,
        A05,A3,A7,A14,
        W05,W3,W7,W14},
    xtick=data,
    xticklabels={
        0.5B,3B,7B,14B,
        0.5B,3B,7B,14B,
        0.5B,3B,7B,14B,
        0.5B,3B,7B,14B},
    x tick label style={font=\scriptsize},
    enlarge x limits=0.04,
    ymajorgrids=true,
    grid style={dashed,gray!25},
    tick pos=left,
    clip=false,
    legend image code/.code={%
        \draw[#1, fill] (0cm,-0.08cm) rectangle (0.18cm,0.08cm);},
    legend style={
        at={(0.5,1.03)}, anchor=south,
        draw=none, legend columns=2, font=\small,
        /tikz/every even column/.append style={column sep=0.18cm}},
]

\addplot+[fill=teal!55, draw=teal!75!black] coordinates {
    (H05,86)  (H3,87)  (H7,86)  (H14,88)
    (F05,74)  (F3,46)  (F7,51)  (F14,46)
    (A05,110) (A3,110) (A7,105) (A14,103)
    (W05,48)  (W3,49)  (W7,49)  (W14,51)};
\addlegendentry{Action tokens (retained)}

\addplot+[fill=gray!30, draw=black!45] coordinates {
    (H05,123) (H3,129) (H7,127) (H14,128)
    (F05,141) (F3,108) (F7,108) (F14,115)
    (A05,223) (A3,217) (A7,213) (A14,208)
    (W05,79)  (W3,92)  (W7,90)  (W14,101)};
\addlegendentry{Thought tokens (removed)}

\node[font=\tiny,color=black!65,rotate=90,anchor=west] at (axis cs:H05,201) {59.1};
\node[font=\tiny,color=black!65,rotate=90,anchor=west] at (axis cs:H3,208)  {59.7};
\node[font=\tiny,color=black!65,rotate=90,anchor=west] at (axis cs:H7,205)  {59.4};
\node[font=\tiny,color=black!65,rotate=90,anchor=west] at (axis cs:H14,208) {59.4};
\node[font=\tiny,color=black!65,rotate=90,anchor=west] at (axis cs:F05,206) {65.6};
\node[font=\tiny,color=black!65,rotate=90,anchor=west] at (axis cs:F3,145)  {70.3};
\node[font=\tiny,color=black!65,rotate=90,anchor=west] at (axis cs:F7,151)  {67.8};
\node[font=\tiny,color=black!65,rotate=90,anchor=west] at (axis cs:F14,153) {71.5};
\node[font=\tiny,color=black!65,rotate=90,anchor=west] at (axis cs:A05,325) {66.9};
\node[font=\tiny,color=black!65,rotate=90,anchor=west] at (axis cs:A3,319)  {66.5};
\node[font=\tiny,color=black!65,rotate=90,anchor=west] at (axis cs:A7,310)  {67.0};
\node[font=\tiny,color=black!65,rotate=90,anchor=west] at (axis cs:A14,303) {66.8};
\node[font=\tiny,color=black!65,rotate=90,anchor=west] at (axis cs:W05,119) {62.5};
\node[font=\tiny,color=black!65,rotate=90,anchor=west] at (axis cs:W3,133)  {65.5};
\node[font=\tiny,color=black!65,rotate=90,anchor=west] at (axis cs:W7,131)  {65.0};
\node[font=\tiny,color=black!65,rotate=90,anchor=west] at (axis cs:W14,144) {66.6};
\node[font=\small,anchor=north] at ([yshift=-10pt] $(axis cs:F14,0)!0.5!(axis cs:A05,0)$) {Student size};
\node[font=\small,anchor=north] at ([yshift=-23pt] $(axis cs:H3,0)!0.5!(axis cs:H7,0)$) {HotpotQA};
\node[font=\small,anchor=north] at ([yshift=-23pt] $(axis cs:F3,0)!0.5!(axis cs:F7,0)$) {FEVER};
\node[font=\small,anchor=north] at ([yshift=-23pt] $(axis cs:A3,0)!0.5!(axis cs:A7,0)$) {ALFWorld};
\node[font=\small,anchor=north] at ([yshift=-23pt] $(axis cs:W3,0)!0.5!(axis cs:W7,0)$) {WebShop};

\draw[black!25] ($(axis cs:H14,0)!0.5!(axis cs:F05,0)$) -- ($(axis cs:H14,400)!0.5!(axis cs:F05,400)$);
\draw[black!25] ($(axis cs:F14,0)!0.5!(axis cs:A05,0)$) -- ($(axis cs:F14,400)!0.5!(axis cs:A05,400)$);
\draw[black!25] ($(axis cs:A14,0)!0.5!(axis cs:W05,0)$) -- ($(axis cs:A14,400)!0.5!(axis cs:W05,400)$);

\end{axis}
\end{tikzpicture}
\vspace{-0.15cm}
\caption{Average assistant-generated tokens per trajectory across high-disclosure settings and student scales, split into action and thought tokens. Numbers above bars show the thought-token share (\%).}
\label{fig:token_reduction}
\end{minipage}%
\hfill
% ============================================================
% (b) Wall-clock reduction
% ============================================================
\begin{minipage}[t]{0.34\textwidth}
\centering
\definecolor{dust}{HTML}{6F8FB7}     % dusty blue
\definecolor{rose}{HTML}{C58A93}      % muted rose
\begin{tikzpicture}[baseline=(wcax.south)]
\begin{axis}[
    name=wcax,
    xbar,
    bar width=6pt,
    scale only axis,
    width=0.68\linewidth,
    height=3cm,
    xlabel={Wall-clock time reduction},
    xlabel style={xshift=-5pt},
    xmin=0, xmax=45,
    xtick={0,10,20,30,40},
    symbolic y coords={WebShop,ALFWorld,FEVER,HotpotQA},
    ytick=data,
    tick label style={font=\small},
    enlarge y limits=0.22,
    xmajorgrids=true,
    grid style={dashed,gray!25},
    tick pos=left,
    clip=false,
    legend image code/.code={%
        \draw[#1, fill] (0cm,-0.07cm) rectangle (0.16cm,0.07cm);},
    legend style={at={(0.5,1.03)}, anchor=south, draw=none,
                  legend columns=2, font=\small,
                  /tikz/every even column/.append style={column sep=0.25cm}},
    nodes near coords={%
        \pgfmathprintnumber[fixed,precision=2]{\pgfplotspointmeta}\%},
    nodes near coords style={font=\scriptsize,anchor=west,xshift=5pt},
    error bars/x dir=both,
    error bars/x explicit,
    error bars/error bar style={line width=3pt},
]

\addplot[fill=LightCyan4, draw=blue!70!black, bar shift=3.5pt,
    error bars/error bar style={line width=0.6pt, black},
] coordinates {
    (5.3,WebShop)    +- (0.91,0)
    (10.2,ALFWorld)  +- (1.06,0)
    (6.3,FEVER)      +- (1.08,0)
    (7.1,HotpotQA)   +- (2.48,0)};
\addlegendentry{Training}

\addplot[fill=DarkSeaGreen3, draw=SlateGray4, bar shift=-3.5pt,
    error bars/error bar style={line width=0.6pt, black},
    coordinate style/.condition={\coordindex==1}{anchor=east, xshift=-36pt},
] coordinates {
    (15.8,WebShop)   +- (3.32,0)
    (41.7,ALFWorld)  +- (2.30,0)
    (11.1,FEVER)     +- (0.62,0)
    (28.4,HotpotQA)  +- (0.56,0)};
\addlegendentry{Inference}
\end{axis}
\end{tikzpicture}
\vspace{0.25cm}
\caption{Wall-clock time reduction (\%) from action-only distillation relative to
full-trajectory distillation.}
\label{fig:wallclock_reduction}
\end{minipage}
\vspace{-0.02cm}
\end{figure*}

\section{Conclusion}
\label{sec:discussion}

We revisit the role of thought supervision in LLM agent distillation by disentangling thought from action supervision within complete trajectories. Across benchmarks, model scales, teacher models, student families, and distillation objectives, we find that thoughts provide little additional benefit when their task-relevant information is already exposed in the teacher's actions. In contrast, removing thoughts becomes harmful when information conveyed by the reasoning is not disclosed in the action. These results collectively suggest that the value of thought supervision depends on what information is absent from the action channel. More broadly, our findings separate the role of reasoning in generating successful actions from its role as supervision for capability transfer. A teacher may need to reason extensively to produce an action, but the student may not need to faithfully imitate that reasoning if the information is already disclosed in the action. This motivates a selective principle for agent distillation: the benefit of thought supervision depends on how much of the task-relevant information in the thought is already exposed in the demonstrated action.

\bibliography{iclr2027_conference}

@inproceedings{chen2021geoqa,
  title={Geoqa: A geometric question answering benchmark towards multimodal numerical reasoning},
  author={Chen, Jiaqi and Tang, Jianheng and Qin, Jinghui and Liang, Xiaodan and Liu, Lingbo and Xing, Eric and Lin, Liang},
  booktitle={Findings of the Association for Computational Linguistics: ACL-IJCNLP 2021},
  pages={513--523},
  year={2021}
}

@inproceedings{ho2020constructing,
  title={Constructing a multi-hop qa dataset for comprehensive evaluation of reasoning steps},
  author={Ho, Xanh and Nguyen, Anh-Khoa Duong and Sugawara, Saku and Aizawa, Akiko},
  booktitle={Proceedings of the 28th International Conference on Computational Linguistics},
  pages={6609--6625},
  year={2020}
}

@article{yang2023intercode,
  title={Intercode: Standardizing and benchmarking interactive coding with execution feedback},
  author={Yang, John and Prabhakar, Akshara and Narasimhan, Karthik and Yao, Shunyu},
  journal={Advances in Neural Information Processing Systems},
  volume={36},
  pages={23826--23854},
  year={2023}
}

@inproceedings{press2023measuring,
  title={Measuring and narrowing the compositionality gap in language models},
  author={Press, Ofir and Zhang, Muru and Min, Sewon and Schmidt, Ludwig and Smith, Noah A and Lewis, Mike},
  booktitle={Findings of the Association for Computational Linguistics: EMNLP 2023},
  pages={5687--5711},
  year={2023}
}

@article{trivedi2022musique,
  title={Musique: Multihop questions via single-hop question composition},
  author={Trivedi, Harsh and Balasubramanian, Niranjan and Khot, Tushar and Sabharwal, Ashish},
  journal={Transactions of the Association for Computational Linguistics},
  volume={10},
  pages={539--554},
  year={2022},
  publisher={MIT Press One Broadway, 12th Floor, Cambridge, Massachusetts 02142, USA~…}
}

@inproceedings{chen2022unigeo,
  title={Unigeo: Unifying geometry logical reasoning via reformulating mathematical expression},
  author={Chen, Jiaqi and Li, Tong and Qin, Jinghui and Lu, Pan and Lin, Liang and Chen, Chongyu and Liang, Xiaodan},
  booktitle={Proceedings of the 2022 conference on empirical methods in natural language processing},
  pages={3313--3323},
  year={2022}
}

@misc{roucher2025smolagents,
  author       = {Aymeric Roucher and Albert Villanova del Moral and Thomas Wolf and Leandro von Werra and Erik Kaunism{\"a}ki},
  title        = {{smolagents}: A smol library to build great agentic systems},
  year         = {2025},
  url = {https://github.com/huggingface/smolagents},
}

@inproceedings{wang2024mint,
  title={Mint: Evaluating llms in multi-turn interaction with tools and language feedback},
  author={Wang, Xingyao and Wang, Zihan and Liu, Jiateng and Chen, Yangyi and Yuan, Lifan and Peng, Hao and Ji, Heng},
  booktitle={International Conference on Learning Representations},
  volume={2024},
  pages={32593--32627},
  year={2024}
}

@article{wang2024executable,
  title={Executable code actions elicit better llm agents},
  author={Wang, Xingyao and Chen, Yangyi and Yuan, Lifan and Zhang, Yizhe and Li, Yunzhu and Peng, Hao and Ji, Heng},
  journal={arXiv preprint arXiv:2402.01030},
  year={2024}
}

@article{jin2025search,
  title={Search-r1: Training llms to reason and leverage search engines with reinforcement learning},
  author={Jin, Bowen and Zeng, Hansi and Yue, Zhenrui and Yoon, Jinsung and Arik, Sercan and Wang, Dong and Zamani, Hamed and Han, Jiawei},
  journal={arXiv preprint arXiv:2503.09516},
  year={2025}
}

@article{wang2022text,
  title={Text embeddings by weakly-supervised contrastive pre-training},
  author={Wang, Liang and Yang, Nan and Huang, Xiaolong and Jiao, Binxing and Yang, Linjun and Jiang, Daxin and Majumder, Rangan and Wei, Furu},
  journal={arXiv preprint arXiv:2212.03533},
  year={2022}
}

@article{chen2023fireact,
  title={Fireact: Toward language agent fine-tuning},
  author={Chen, Baian and Shu, Chang and Shareghi, Ehsan and Collier, Nigel and Narasimhan, Karthik and Yao, Shunyu},
  journal={arXiv preprint arXiv:2310.05915},
  year={2023}
}

@inproceedings{zeng2024agenttuning,
  title={Agenttuning: Enabling generalized agent abilities for llms},
  author={Zeng, Aohan and Liu, Mingdao and Lu, Rui and Wang, Bowen and Liu, Xiao and Dong, Yuxiao and Tang, Jie},
  booktitle={Findings of the Association for Computational Linguistics: ACL 2024},
  pages={3053--3077},
  year={2024}
}

@inproceedings{yang2018hotpotqa,
  title={HotpotQA: A dataset for diverse, explainable multi-hop question answering},
  author={Yang, Zhilin and Qi, Peng and Zhang, Saizheng and Bengio, Yoshua and Cohen, William and Salakhutdinov, Ruslan and Manning, Christopher D},
  booktitle={Proceedings of the 2018 conference on empirical methods in natural language processing},
  pages={2369--2380},
  year={2018}
}

@inproceedings{thorne2018fever,
  title={FEVER: a large-scale dataset for fact extraction and VERification},
  author={Thorne, James and Vlachos, Andreas and Christodoulopoulos, Christos and Mittal, Arpit},
  booktitle={Proceedings of the 2018 Conference of the North American Chapter of the Association for Computational Linguistics: Human Language Technologies, Volume 1 (Long Papers)},
  pages={809--819},
  year={2018}
}

@inproceedings{shridhar2021alfworld,
title={{{ALFW}}orld: Aligning Text and Embodied Environments for Interactive Learning},
author={Mohit Shridhar and Xingdi Yuan and Marc-Alexandre Cote and Yonatan Bisk and Adam Trischler and Matthew Hausknecht},
booktitle={International Conference on Learning Representations},
year={2021},
}

@article{yao2022webshop,
  title={Webshop: Towards scalable real-world web interaction with grounded language agents},
  author={Yao, Shunyu and Chen, Howard and Yang, John and Narasimhan, Karthik},
  journal={Advances in Neural Information Processing Systems},
  volume={35},
  pages={20744--20757},
  year={2022}
}

@inproceedings{yao2023react,
title={ReAct: Synergizing Reasoning and Acting in Language Models},
author={Shunyu Yao and Jeffrey Zhao and Dian Yu and Nan Du and Izhak Shafran and Karthik R Narasimhan and Yuan Cao},
booktitle={The Eleventh International Conference on Learning Representations },
year={2023},
}

@inproceedings{liu2026structured,
  title={Structured Agent Distillation for Large Language Model Agents},
  author={Liu, Jun and Kong, Zhenglun and Dong, Peiyan and Yang, Changdi and Li, Tianqin and Xie, Yanyue and Gong, Yifan and Shen, Xuan and Zhao, Pu and Tang, Hao and others},
  booktitle={Proc. of the 25th International Conference on Autonomous Agents and Multiagent Systems},
  pages={3676--3685},
  year={2026}
}

@article{qiu2025agentdistill,
  title={Agentdistill: Training-free agent distillation with generalizable mcp boxes},
  author={Qiu, Jiahao and Juan, Xinzhe and Wang, Yimin and Yang, Ling and Qi, Xuan and Zhang, Tongcheng and Guo, Jiacheng and Lu, Yifu and Yao, Zixin and Wang, Hongru and others},
  journal={arXiv preprint arXiv:2506.14728},
  year={2025}
}

@article{kang2026distilling,
  title={Distilling llm agent into small models with retrieval and code tools},
  author={Kang, Minki and Jeong, Jongwon and Lee, Seanie and Cho, Jaewoong and Hwang, Sung Ju},
  journal={Advances in Neural Information Processing Systems},
  volume={38},
  pages={106501--106538},
  year={2026}
}

@article{yang2024qwen25,
  title={Qwen2.5 Technical Report},
  author={Yang, An and Yang, Baosong and Zhang, Beichen and Hui, Binyuan
          and Zheng, Bo and Yu, Bowen and others},
  journal={arXiv preprint arXiv:2412.15115},
  year={2024}
}

@inproceedings{chen2021finqa,
  title={Finqa: A dataset of numerical reasoning over financial data},
  author={Chen, Zhiyu and Chen, Wenhu and Smiley, Charese and Shah, Sameena and Borova, Iana and Langdon, Dylan and Moussa, Reema and Beane, Matt and Huang, Ting-Hao and Routledge, Bryan R and others},
  booktitle={Proceedings of the 2021 Conference on Empirical Methods in Natural Language Processing},
  pages={3697--3711},
  year={2021}
}

@inproceedings{zhao2022multihiertt,
  title={MultiHiertt: Numerical reasoning over multi hierarchical tabular and textual data},
  author={Zhao, Yilun and Li, Yunxiang and Li, Chenying and Zhang, Rui},
  booktitle={Proceedings of the 60th Annual Meeting of the Association for Computational Linguistics (Volume 1: Long Papers)},
  pages={6588--6600},
  year={2022}
}

@inproceedings{amini2019mathqa,
  title={Mathqa: Towards interpretable math word problem solving with operation-based formalisms},
  author={Amini, Aida and Gabriel, Saadia and Lin, Shanchuan and Koncel-Kedziorski, Rik and Choi, Yejin and Hajishirzi, Hannaneh},
  booktitle={Proceedings of the 2019 Conference of the North American Chapter of the Association for Computational Linguistics: Human Language Technologies, Volume 1 (Long and Short Papers)},
  pages={2357--2367},
  year={2019}
}

@misc{openai2026gpt54,
  author       = {{OpenAI}},
  title        = {Introducing {GPT-5.4}},
  year         = {2026},
  month        = mar,
  howpublished = {\url{https://openai.com/index/introducing-gpt-5-4/}}
}

@misc{anthropic2026opus5,
  author       = {{Anthropic}},
  title        = {Introducing {Claude Opus 5}},
  year         = {2026},
  month        = jul,
  howpublished = {\url{https://www.anthropic.com/news/claude-opus-5}}
}

@misc{google2026gemini31,
  author       = {{Google}},
  title        = {{Gemini 3.1 Pro}: A smarter model for your most complex tasks},
  year         = {2026},
  month        = feb,
  howpublished = {\url{https://blog.google/innovation-and-ai/models-and-research/gemini-models/gemini-3-1-pro/}}
}

@misc{meta2024llama32,
  author       = {{Meta AI}},
  title        = {{Llama 3.2}: Revolutionizing Edge AI and Vision with Open, Customizable Models},
  year         = {2024},
  month        = sep,
  howpublished = {\url{https://ai.meta.com/blog/llama-3-2-connect-2024-vision-edge-mobile-devices/}}
}

@inproceedings{penaloza2026privileged,
title={Privileged Information Distillation for Language Models},
author={Emiliano Penaloza and Dheeraj Vattikonda and Nicolas Gontier and Alexandre Lacoste and Laurent Charlin and Massimo Caccia},
booktitle={Forty-third International Conference on Machine Learning},
year={2026}
}

@inproceedings{li2023symbolic,
  title={Symbolic chain-of-thought distillation: Small models can also “think” step-by-step},
  author={Li, Liunian Harold and Hessel, Jack and Yu, Youngjae and Ren, Xiang and Chang, Kai-Wei and Choi, Yejin},
  booktitle={Proceedings of the 61st Annual Meeting of the Association for Computational Linguistics (Volume 1: Long Papers)},
  pages={2665--2679},
  year={2023}
}

@inproceedings{hashemzadeh2024subgoal,
  title = {Sub-goal Distillation: A Method to Improve Small Language Agents},
  author = {Hashemzadeh, Maryam and Stengel-Eskin, Elias and Chandar, Sarath and C{\^{o}}t{\'{e}}, Marc-Alexandre},
  booktitle = {Proceedings of The 3rd Conference on Lifelong Learning Agents},
  pages = {1053--1075},
  year = {2025},
  volume = {274},
  series = {Proceedings of Machine Learning Research},
  publisher = {PMLR},
}

@inproceedings{chen2024agent,
  title={Agent-flan: Designing data and methods of effective agent tuning for large language models},
  author={Chen, Zehui and Liu, Kuikun and Wang, Qiuchen and Zhang, Wenwei and Liu, Jiangning and Lin, Dahua and Chen, Kai and Zhao, Feng},
  booktitle={Findings of the Association for Computational Linguistics: ACL 2024},
  pages={9354--9366},
  year={2024}
}

@misc{openai2024gpt4omini,
  author={OpenAI},
  title={{GPT-4o mini: Advancing cost-efficient intelligence}},
  year={2024},
  note={\url{https://openai.com/index/gpt-4o-mini-advancing-cost-efficient-intelligence}}
}

@article{liu2026cot,
  title={Where Do CoT Training Gains Land in LLM based Agents?},
  author={Liu, Jingyu and Wang, Zhiwen and Jing, Yuxin and Zhou, Huanyu and Liu, Yong},
  journal={arXiv preprint arXiv:2606.26935},
  year={2026}
}

@inproceedings{wadhwa2024investigating,
  title={Investigating mysteries of cot-augmented distillation},
  author={Wadhwa, Somin and Amir, Silvio and Wallace, Byron C},
  booktitle={Proceedings of the 2024 Conference on Empirical Methods in Natural Language Processing},
  pages={6071--6086},
  year={2024}
}

@inproceedings{chen2025unveiling,
  title={Unveiling the key factors for distilling chain-of-thought reasoning},
  author={Chen, Xinghao and Sun, Zhijing and Wenjin, Guo and Zhang, Miaoran and Chen, Yanjun and Sun, Yirong and Su, Hui and Pan, Yijie and Klakow, Dietrich and Li, Wenjie and others},
  booktitle={Findings of the association for computational linguistics: ACL 2025},
  pages={15094--15119},
  year={2025}
}

@inproceedings{hsieh2023distilling,
  title={Distilling step-by-step! outperforming larger language models with less training data and smaller model sizes},
  author={Hsieh, Cheng-Yu and Li, Chun-Liang and Yeh, Chih-Kuan and Nakhost, Hootan and Fujii, Yasuhisa and Ratner, Alex and Krishna, Ranjay and Lee, Chen-Yu and Pfister, Tomas},
  booktitle={Findings of the association for computational linguistics: ACL 2023},
  pages={8003--8017},
  year={2023}
}

@inproceedings{magister2023teaching,
  title={Teaching small language models to reason},
  author={Magister, Lucie Charlotte and Mallinson, Jonathan and Adamek, Jakub and Malmi, Eric and Severyn, Aliaksei},
  booktitle={Proceedings of the 61st Annual Meeting of the Association for Computational Linguistics (Volume 2: Short Papers)},
  pages={1773--1781},
  year={2023}
}

@article{abouelenin2025phi,
  title={Phi-4-mini technical report: Compact yet powerful multimodal language models via mixture-of-loras},
  author={Abouelenin, Abdelrahman and Ashfaq, Atabak and Atkinson, Adam and Awadalla, Hany and Bach, Nguyen and Bao, Jianmin and Benhaim, Alon and Cai, Martin and Chaudhary, Vishrav and Chen, Congcong and others},
  journal={arXiv preprint arXiv:2503.01743},
  year={2025}
}

@article{lyu2025correction,
  title={From correction to mastery: Reinforced distillation of large language model agents},
  author={Lyu, Yuanjie and Wang, Chengyu and Huang, Jun and Xu, Tong},
  journal={arXiv preprint arXiv:2509.14257},
  year={2025}
}

@inproceedings{tang2026smartad,
  title={SmartAD: Capacity-Aligned Agent Distillation for Small Language Models},
  author={Tang, Guokai and Zhao, Feng},
  booktitle={Findings of the Association for Computational Linguistics: ACL 2026},
  pages={27045--27057},
  year={2026}
}
\bibliographystyle{iclr2027_conference}

\clearpage
\appendix

\section{Datasets and Trajectory Construction}
\label{app:data}

Table~\ref{tab:dataset_statistics} summarizes the datasets used in our agent distillation experiments. We consider eight benchmarks covering multi-hop question answering (HotpotQA~\citep{yang2018hotpotqa}), fact verification (FEVER~\citep{thorne2018fever}), interactive tasks (ALFWorld~\citep{shridhar2021alfworld} and WebShop~\citep{yao2022webshop}), and quantitative reasoning (FinQA~\citep{chen2021finqa}, MultiHiertt~\citep{zhao2022multihiertt}, MathQA~\citep{amini2019mathqa}, and GeoQA~\citep{chen2021geoqa}). HotpotQA~\citep{yang2018hotpotqa} is a knowledge-intensive multi-hop question answering benchmark requiring iterative retrieval and reasoning over Wikipedia. FEVER~\citep{thorne2018fever} is fact verification, where the agent retrieves evidence and classifies a claim as supported, refuted, or unverifiable. WebShop~\citep{yao2022webshop} is for web navigation, where the agent searches, compares, and purchases a product matching a natural language instruction. For embodied decision-making, ALFWorld~\citep{shridhar2021alfworld} is a text-based household environment requiring long-horizon planning. 

For quantitative reasoning, FinQA~\citep{chen2021finqa} is a financial question answering benchmark built on earnings reports, where the agent must extract relevant values from both text and tables and perform multi-step numerical computation. MultiHiertt~\citep{zhao2022multihiertt} extends this setting to documents with multiple hierarchical tables and long unstructured text, requiring the agent to locate evidence across several tables before carrying out numerical reasoning. MathQA~\citep{amini2019mathqa} is a large-scale collection of multiple-choice math word problems annotated with operation programs, testing the agent's ability to translate natural language problems into correct sequences of arithmetic operations. GeoQA~\citep{chen2021geoqa} is a geometry problem-solving benchmark collected from middle school exams, where the agent must reason about geometric relations such as angles, lengths, and areas to compute the answer; we use its English version released with UniGeo~\citep{chen2022unigeo} and train and evaluate on a text-only subset of problems solvable without the diagram. %For each benchmark, \emph{Teacher Trajectories} reports the size of the teacher trajectory pool, \emph{Training Data} reports the number of trajectories used for student training using Qwen2.5-32B-Instruct, and \emph{Eval Set Size} reports the number of evaluation instances. For HotpotQA, FEVER, 2WikiMultiHopQA, and MuSiQue, we use at most 500 evaluation instances, following the evaluation protocol of prior agent distillation work~\citep{kang2026distilling, wang2024mint}. For ALFWorld, we evaluate on the full validation set of 274 games, comprising 140 \texttt{valid\_seen} and 134 \texttt{valid\_unseen} games.

Unless otherwise specified, teacher trajectories are generated using Qwen2.5-32B-Instruct~\citep{yang2024qwen25} in the ReAct~\citep{yao2023react} format. For the quantitative reasoning benchmarks, actions are instead constructed from dataset-provided symbolic programs, as described in Section~\ref{sec:high-vs-low}. Only trajectories that satisfy the benchmark-specific success criterion are eligible for distillation; the table reports the final training set sizes after selection. During training, thoughts and actions serve as supervised targets, while observations provide context and are masked out from the loss. For each benchmark, the clean baseline and all corruption conditions use the same underlying training trajectories and evaluation set. 

For the out-of-distribution evaluation in Section~\ref{sec:ood}, we additionally consider 2WikiMultiHopQA~\citep{ho2020constructing}, Bamboogle~\citep{press2023measuring}, and MuSiQue~\citep{trivedi2022musique}. These benchmarks are used only for evaluation, without target-dataset training or fine-tuning.

For each benchmark, \emph{Teacher Trajectories} reports the size of the teacher trajectory pool, \emph{Training Data} reports the number of trajectories used for student training using Qwen2.5-32B-Instruct, and \emph{Eval Set Size} reports the number of evaluation instances. For HotpotQA, FEVER, 2WikiMultiHopQA, and MuSiQue, we use at most 500 evaluation instances, following the evaluation protocol of prior agent distillation work~\citep{kang2026distilling, wang2024mint}. For ALFWorld, we evaluate on the full validation set of 274 games, comprising 140 \texttt{valid\_seen} and 134 \texttt{valid\_unseen} games.

\begin{table}[h]
\centering
\caption{Dataset statistics for agent distillation experiments.}
\label{tab:dataset_statistics}
\begin{tabular}{lccc}
\toprule
Benchmark & Teacher Trajectories & Training Data & Eval Set Size \\
\midrule
HotpotQA & 1,000 & 597 & 500 \\
FEVER    & 1,000 & 650 & 500 \\
ALFWorld & 3,553 & 2,872 & 274 \\
WebShop  & 10,587 & 3,555  & 500  \\
\midrule
FinQA        & 5,695  & 1,500 & 300 \\
MultiHiertt  & 6,217  & 1,500 & 832 \\
MathQA       & 17,167 & 1,500 & 1,000 \\
GeoQA        & 2,362  & 1,500   & 500 \\
% MATH         & 1,000  & 601   & 500 \\
\midrule
2WikiMultiHopQA & -- & -- & 500 \\
Bamboogle & -- & -- & 125 \\
MuSiQue & -- & -- & 500 \\
\bottomrule
\end{tabular}
\end{table}

\section{Training, Inference, and Evaluation Details}
\label{app:eval-detail}

\subsection{Training}

\paragraph{KL Loss Function.}
We evaluate KL-based distillation as an alternative to the SFT objective. For each training trajectory, we minimize the token-level forward KL divergence between the teacher and student output distributions:
\begin{equation}
\mathcal{L}_{\text{KL}}(\theta) = \mathbb{E}_{\tau \sim \mathcal{D}}
\left[
\sum_{j \in \mathcal{S}(\tau)} \mathrm{KL}\!\left( \pi_{\text{teacher}}(\cdot \mid y_{<j}) 
\,\|\, \pi_\theta(\cdot \mid y_{<j}) \right)
\right]
\label{eq:kl-loss}
\end{equation}
Both distributions are conditioned on the same trajectory prefix $y_{<j}$. The set $S(\tau)$ contains the supervised token positions under the corresponding intervention, following the same masking convention as the SFT objective.

\paragraph{Retrieval Environment.}
For reproducibility of experiments, we use a fixed 2018 Wikipedia corpus as the retrieval knowledge base. Following Search-R1~\citep{jin2025search}, we encode both documents and search queries using \texttt{e5-base-v2}~\citep{wang2022text} and retrieve the top-$3$ passages using an exact flat index. The resulting retriever is exposed to the agent as a \texttt{web\_search} tool. The agent is implemented following the CodeAct~\citep{wang2024executable} framework in \texttt{smolagents}~\citep{roucher2025smolagents}. We use the same Wikipedia corpus and retriever across all retrieval-based conditions to ensure a consistent external knowledge source.

\subsection{Inference}
\paragraph{Inference Prompt.}
The inference prompt presented to the student is assembled from a system prompt specifying the agent interaction format, a user prompt provided by the agent framework or benchmark, and the benchmark-specific task input (Figures~\ref{fig:inference-prompt}, \ref{fig:inference-prompt-alfworld}, and \ref{fig:inference-prompt-webshop}). For HotpotQA, FEVER, FinQA, MultiHiertt, MathQA, and GeoQA, we use a shared ReAct-style system prompt that defines the Thought--Code--Observation interaction cycle and available tools. ALFWorld and WebShop instead use benchmark-specific interaction prompts that define their respective action spaces and include fixed demonstration examples. For MathQA, we omit the multiple-choice options and evaluate it in an open-ended setting. Within each benchmark, the prompt template is held fixed across all intervention conditions, and only the task instance varies.

\begin{figure}[ht]
\begin{lstlisting}[style=promptbox]
=== System prompt ===
You are an expert assistant who can solve any task using code blobs. You will be given a task to solve as best 
you can.
To do so, you have been given access to a list of tools: these tools are basically Python functions which you 
can call with code.
To solve the task, you must plan forward to proceed in a series of steps, in a cycle of 'Thought:', 'Code:', 
and 'Observation:' sequences.
 
At each step, in the 'Thought:' sequence, you should first explain your reasoning towards solving the task and 
the tools that you want to use.
Then in the 'Code:' sequence, you should write the code in simple Python. The code sequence must start with 
'```py' and end with '```<end_code>', else you will fail. For example:
Code:
```py
result = 2 + 2
final_answer(result)
```<end_code>
During each intermediate step, you can use 'print()' to save whatever important information you will then need.
These print outputs will then appear in the 'Observation:' field, which will be available as input for the next 
step.
In the end you have to return a final answer using the `final_answer` tool.
 
On top of performing computations in the Python code snippets that you create, you only have access to these 
tools:
- web_search: Uses semantic search to retrieve the parts of 2018 wikipedia that could be most relevant to 
answer your query.
    Takes inputs: {'query': {'type': 'string', 'description': '...'}}
    Returns an output of type: string
 
Now Begin! If you solve the task correctly, you will receive a reward of $1,000,000.
 
=== User prompt ===
New task:
{task}
 
=== Benchmark-specific task input ===
HotpotQA / FEVER   : {question or claim}
FinQA / MultiHiertt : {question} + "\n\nContext:\n" + {pre_text, table, post_text}
MathQA / GeoQA : {question}
\end{lstlisting}
\caption{Assembled inference prompt for HotpotQA, FEVER, FinQA, MultiHiertt, MathQA, and GeoQA.}
\label{fig:inference-prompt}
\end{figure}

\begin{figure}[ht]
\begin{lstlisting}[style=promptbox]
=== System prompt ===
You are an agent interacting with the ALFWorld household environment.
At each turn, output exactly one command and nothing else. Do not add a prefix such as "Action:" or 
"Assistant:", and do not explain an environment command outside a `think:` command.

Valid command forms are:
- think: <brief reasoning about the next action>
- go to <receptacle>
- open <receptacle>
- close <receptacle>
- take <object> from <receptacle>
- put <object> in/on <receptacle>
- use <object>
- heat <object> with <receptacle>
- clean <object> with <receptacle>
- cool <object> with <receptacle>
- inventory
- look
- examine <object or receptacle>

Use the exact object and receptacle names and numeric IDs shown by the environment. The `think:` command does 
not change the environment. Never invent observations; wait for the environment response after each command.

=== User prompt ===
Interact with a household to solve a task. Here are two examples.
{Example 1} + {Example 2}

Here is the task.
{room description} + "\nYour task is to: " + {task instruction}

\end{lstlisting}
\caption{Assembled inference prompt for ALFWorld.}
\label{fig:inference-prompt-alfworld}
\end{figure}

 \begin{figure}[ht]
 \begin{lstlisting}[style=promptbox]
=== System prompt ===
You are an agent shopping on the WebShop website to fulfill a customer instruction. At each turn, output 
exactly one command and nothing else. Do not add a prefix such as "Action:" or "Assistant:", and do not explain
a website command outside a `think[...]` command.

Valid command forms are:
- think[<brief reasoning about the next action>]
- search[<query>]
- click[<exact text of a button, product code, or option shown on the page>]

`search[...]` is only available on the search page. `click[...]` must use the exact text shown in brackets on 
the current page. The `think[...]` command does not change the page. Buying with click[Buy Now] ends the 
episode, so select the required options first. Never invent observations; wait for the page after each command.

=== User prompt ===
{Example 1}

Here is the task.
{webshop_instruction}
{initial_search_page}

\end{lstlisting}
\caption{Assembled inference prompt for WebShop.}
\label{fig:inference-prompt-webshop}
\end{figure}

\paragraph{Decoding.}
For ALFWorld and WebShop, each decoding call generates a single-line command with a 100-token generation limit. Episodes are capped at 49 and 15 environment-action steps, respectively. The \texttt{think:} and \texttt{think[...]} commands do not count toward these interaction budgets and do not consume an environment-action step. Thus, students trained with and without thoughts are allowed the same maximum number of environment-interacting actions. For all other benchmarks, each agent step jointly generates the thought and code with a limit of $1{,}024$ new tokens, and episodes are capped at five steps. The fallback that forces a terminal answer when the interaction budget is exhausted is disabled, so any episode that exceeds the budget is scored as a failure.

\subsection{Evaluation}
\paragraph{Success Criteria.}
An LLM judge is used for HotpotQA, 2WikiMultiHopQA, Bamboogle, and MuSiQue. We use \texttt{GPT-4o-mini} with structured decoding that constrains the output to a JSON object containing a free-text explanation and a binary score; Figure~\ref{fig:judge-prompt} gives the prompt verbatim. FEVER predictions are one of three labels and are scored by matching the normalized prediction against the gold label. FinQA, MultiHiertt, MathQA, and GeoQA are scored by symbolic and numeric equivalence on the extracted final answer, so that equivalent surface forms are treated identically. WebShop and ALFWorld use the benchmark-provided success signal.

\begin{figure}[ht]
\begin{lstlisting}[style=promptbox]
Compare the following predicted answer with the gold (correct) answer and determine if they match in meaning 
given the question.
Be reasonable - phrasing differences are okay if the core meaning is the same. You should accept the aliases 
and the answer conveying the same conclusion.
If the predicted answer is overly verbose and fails to capture the key information found in the gold answer 
(i.e., low recall), consider it a false answer.
 
Question: {question}
Predicted answer: {predicted}
Gold answer: {gold}
 
Output your evaluation as a JSON object with two fields:
- explanation: Brief explanation of your scoring decision
- score: 0 or 1 indicating if answers match
\end{lstlisting}
\caption{Judge prompt used for HotpotQA, 2WikiMultiHopQA, Bamboogle, and MuSiQue.}
\label{fig:judge-prompt}
\end{figure}

\section{Corruption Operator Implementation}
\label{app:corruption}

\paragraph{Paraphrase.}
The \texttt{Paraphrase} operator is instantiated with Qwen2.5-32B-Instruct, the same model used as our default teacher. Each selected thought is rewritten independently: the paraphraser receives the body of a single thought $t_i$, and its output is reinserted in place of $t_i$, so that only the surface form of the thought changes. The prompt (Figure~\ref{fig:para_prompt}) instructs the model to diverge from the original in wording and sentence structure as much as possible while holding every fact, entity, number, plan step, and conclusion fixed, and to preserve approximate length. We decode with temperature $0.8$, top-$p$ $0.95$, a $512$-token cap, and a fixed seed ($0$). Paraphrased thoughts remain supervised targets and are not masked from the loss.

\begin{figure}[ht]
\begin{lstlisting}[basicstyle=\ttfamily\scriptsize, breaklines=true, frame=single]
=== System prompt ===
You rewrite one step of an AI agent's reasoning. Rewrite the user's text so that the wording and sentence 
structure are as different as possible while the meaning stays exactly the same: keep every fact, entity, 
number, plan step, and conclusion; do not add, remove, or correct anything. Keep roughly the same length. 
Output only the rewritten text, with no prefix, no quotes, and no commentary.

=== User prompt ===
{thought}
\end{lstlisting}
\caption{Prompt used for the \texttt{Paraphrase} intervention (Eq.~\ref{eq:corruption-ops}). The placeholder \texttt{\{thought\}} is replaced by the natural-language body of a single teacher thought $t_i$.}
\label{fig:para_prompt}
\end{figure}

\paragraph{Replace.}
For \texttt{Replace}, each selected thought is independently substituted with a thought sampled uniformly at random from the pool of thoughts of other training instances of the same benchmark.

\paragraph{Drop.}
For \texttt{Drop}, the selected thought is removed from the trajectory entirely, together with its span-level formatting markers, rather than being replaced by an empty placeholder or masked from the loss.

\section{Full Intervention Results}
\label{app:exp}

This appendix provides the full numerical results underlying the analyses in the main text, together with detailed corruption-ratio sweeps. All accuracy and success-rate values are reported as percentages.

\paragraph{Thought and action corruption across student scales
(Sections~\ref{sec:action-matters}--\ref{sec:student-scale}).}
Table~\ref{tab:student-size} reports the results of thought and action interventions across four Qwen2.5 student scales: 0.5B, 3B, 7B, and 14B. The 3B results support the comparison between thought and action corruption in Section~\ref{sec:action-matters}, while the thought-intervention results across all four scales support Section~\ref{sec:student-scale}. Each intervention is compared with the clean baseline for the same student model.

\begin{table}[h]
\centering
\setlength{\tabcolsep}{5pt}
\caption{Task accuracy under thought and action corruption across Qwen2.5-\{0.5B, 3B, 7B, 14B\}-Instruct students distilled from Qwen2.5-32B-Instruct ($p{=}1.0$). Results are averaged over three independent training and evaluation runs, with standard deviations reported in parentheses.}
\label{tab:student-size}
\begin{tabular}{llcccccc}
\toprule
& & & \multicolumn{3}{c}{Thought} & \multicolumn{2}{c}{Action} \\
\cmidrule(lr){4-6}\cmidrule(lr){7-8}
Benchmark & Size & Clean & Drop & Replace & Paraphrase & Drop & Replace \\
\midrule
\multirow{4}{*}{HotpotQA}
 & 0.5B & 33.4\std{0.9} & 34.2\std{0.6} & 31.0\std{1.5} & 31.0\std{1.2} & \phantom{0}7.2\std{1.6} & 10.0\std{0.5} \\
 & 3B & 48.0\std{0.7} & 46.8\std{0.3} & 45.6\std{1.5} & 46.8\std{1.3} & 25.5\std{1.4} & \phantom{0}5.0\std{1.0} \\
 & 7B   & 50.8\std{1.1} & 54.0\std{0.6} & 48.0\std{1.7} & 52.2\std{1.6} & 29.0\std{1.8} & \phantom{0}4.2\std{2.1} \\
 & 14B  & 56.4\std{1.2} & 58.6\std{0.9} & 52.4\std{1.8} & 55.0\std{1.2} & \phantom{0}2.2\std{1.3} & \phantom{0}5.2\std{1.7} \\
\midrule
\multirow{4}{*}{FEVER}
 & 0.5B & 51.2\std{0.8} & 48.8\std{0.8} & 52.4\std{1.2} & 50.6\std{0.5} & 40.0\std{1.2} & 44.6\std{1.2} \\
 & 3B   & 63.4\std{1.2} & 64.9\std{0.8} & 63.9\std{2.5} & 62.0\std{1.2} & 50.4\std{1.1} & 29.3\std{6.9} \\
 & 7B   & 66.4\std{1.1} & 67.8\std{0.9} & 68.4\std{1.5} & 65.6\std{1.1} & 50.0\std{1.0} & 25.6\std{4.5} \\
 & 14B  & 69.8\std{1.1} & 65.6\std{1.1} & 66.6\std{1.2} & 66.0\std{0.9} & 51.0\std{1.2} & 18.6\std{3.5} \\
\midrule
\multirow{4}{*}{WebShop}
 & 0.5B & 43.0\std{0.5} & 40.6\std{0.4} & 43.0\std{0.9} & 39.6\std{0.8} & \phantom{0}0.0\std{0.0} & \phantom{0}0.0\std{0.0} \\
 & 3B   & 40.2\std{0.9} & 43.1\std{0.8} & 43.7\std{1.0} & 37.8\std{0.9} & \phantom{0}0.0\std{0.0} & \phantom{0}0.0\std{0.0} \\
 & 7B   & 41.4\std{0.9} & 42.2\std{0.7} & 42.0\std{1.3} & 44.0\std{1.4} & \phantom{0}0.0\std{0.0} & \phantom{0}0.0\std{0.0} \\
 & 14B  & 41.4\std{0.5} & 43.8\std{0.4} & 41.4\std{1.4} & 42.0\std{1.1} & \phantom{0}0.0\std{0.0} & \phantom{0}0.0\std{0.0} \\
\midrule
\multirow{4}{*}{ALFWorld}
 & 0.5B & 88.0\std{2.3} & 96.7\std{1.5} & 94.9\std{2.9} & 87.6\std{2.0} & \phantom{0}0.0\std{0.0} & \phantom{0}0.0\std{0.0} \\
 & 3B   & 88.3\std{1.0} & 95.1\std{0.9} & 94.9\std{1.0} & 90.3\std{1.7} & \phantom{0}0.0\std{0.0} & \phantom{0}0.0\std{0.0} \\
 & 7B   & 90.5\std{1.1} & 95.6\std{1.0} & 96.7\std{1.7} & 90.5\std{1.8} & \phantom{0}0.0\std{0.0} & \phantom{0}0.0\std{0.0} \\
 & 14B  & 92.3\std{1.0} & 97.6\std{0.9} & 97.1\std{0.7} & 92.0\std{1.3} & \phantom{0}0.0\std{0.0} & \phantom{0}0.0\std{0.0} \\
\bottomrule
\end{tabular}
\end{table}

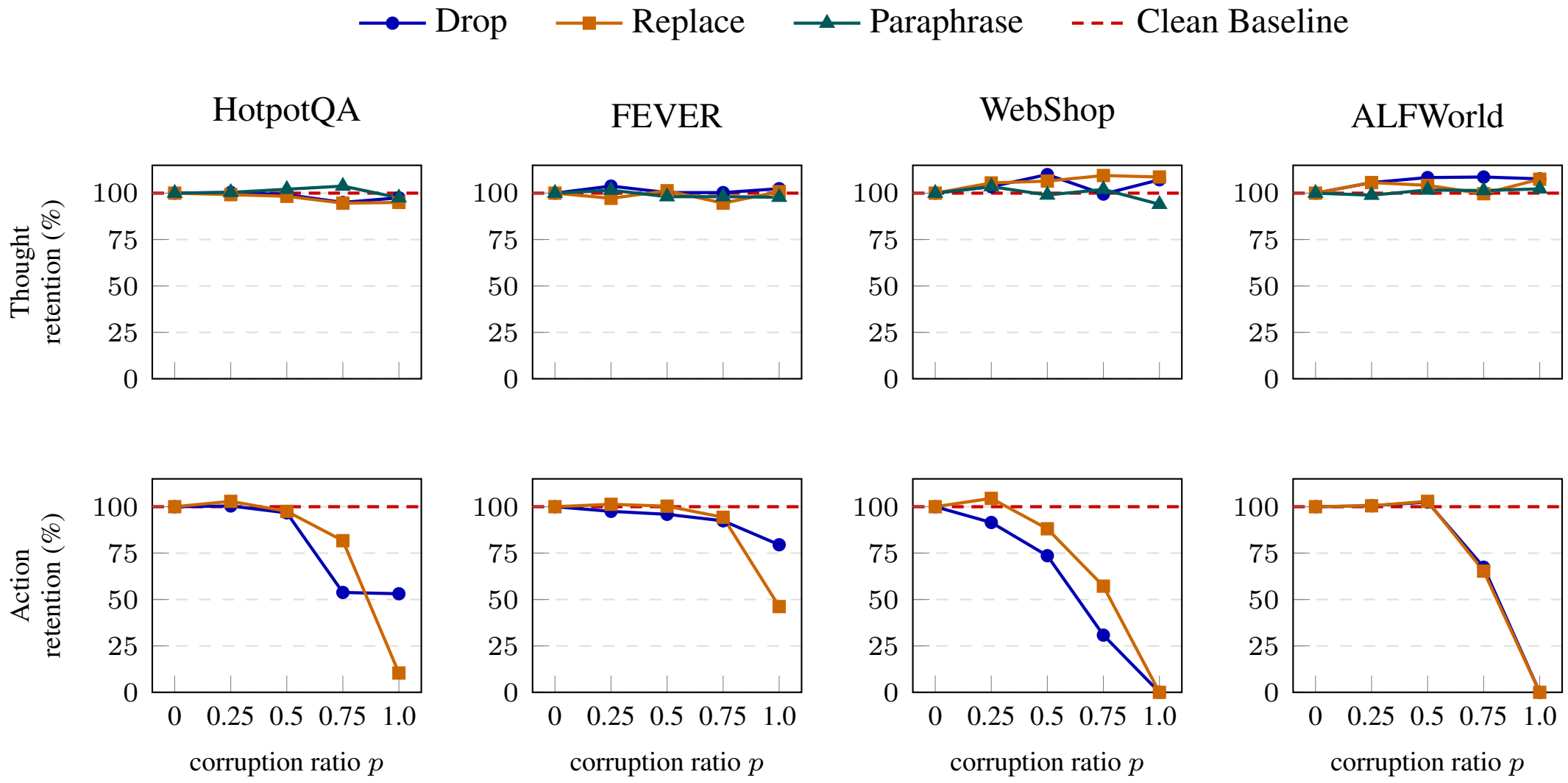
\begin{figure}[ht]
\centering
\begin{tikzpicture}[
  dropline/.style   ={mark=*, mark size=1.3pt, line width=0.8pt, blue!70!black},
  replaceline/.style={mark=square*, mark size=1.3pt, line width=0.8pt, orange!80!black},
  paraline/.style   ={mark=triangle*,mark size=1.6pt, line width=0.8pt, teal!70!black},
]
\begin{groupplot}[
  group style={
    group size=4 by 2,
    horizontal sep=1.0cm,
    vertical sep=0.9cm,
    x descriptions at=edge bottom,
  },
  width=4cm, height=3.5cm,
  xlabel={corruption ratio $p$},
  xtick={0,0.25,0.5,0.75,1}, xticklabels={0,0.25,0.5,0.75,1.0},
  tick pos=left,
  x tick label style={font=\scriptsize},
  y tick label style={font=\scriptsize},
  label style={font=\scriptsize},
  title style={font=\small},
  ymajorgrids=true,
  grid style={dashed,gray!25},
  extra y ticks={100},
  extra y tick labels={},
  extra y tick style={grid=major, major grid style={thick,red!80!black,densely dashed}},
]
 
% ---------------- Row 1: Thought corruption ----------------
\nextgroupplot[
  title={HotpotQA},
  ylabel={Thought\\retention (\%)}, ylabel style={align=center},
  ymin=0, ymax=115, ytick={0,25,50,75,100},
  legend to name=sweeplegend,
  legend columns=-1,
  legend style={draw=none, font=\small,
    /tikz/every even column/.append style={column sep=0.35cm}},
]
\addplot[dropline]    coordinates {(0,100) (0.25,100.4) (0.5,99.2) (0.75,95.0) (1,97.5)};
\addlegendentry{Drop}
\addplot[replaceline] coordinates {(0,100) (0.25,99.2) (0.5,98.3) (0.75,94.6) (1,95.0)};
\addlegendentry{Replace}
\addplot[paraline]    coordinates {(0,100) (0.25,100.4) (0.5,102.1) (0.75,103.8) (1,97.5)};
\addlegendentry{Paraphrase}
\addlegendimage{legend image code/.code={%
  \draw[thick, red!80!black, densely dashed] (0cm,0cm) -- (0.5cm,0cm);}}
\addlegendentry{Clean Baseline}

\nextgroupplot[title={FEVER}, ymin=0, ymax=115, ytick={0,25,50,75,100}]
\addplot[dropline]    coordinates {(0,100) (0.25,103.8) (0.5,100.3) (0.75,100.3) (1,102.4)};
\addplot[replaceline] coordinates {(0,100) (0.25,97.2) (0.5,101.3) (0.75,94.6) (1,100.8)};
\addplot[paraline]    coordinates {(0,100) (0.25,101.6) (0.5,98.1) (0.75,98.1) (1,97.8)};

\nextgroupplot[title={WebShop}, ymin=0, ymax=115, ytick={0,25,50,75,100}]
\addplot[dropline]    coordinates {(0,100) (0.25,103.5) (0.5,110.0) (0.75,99.5) (1,107.2)};
\addplot[replaceline] coordinates {(0,100) (0.25,105.5) (0.5,106.5) (0.75,109.5) (1,108.7)};
\addplot[paraline]    coordinates {(0,100) (0.25,103.5) (0.5,99.0) (0.75,102.0) (1,94.0)};

\nextgroupplot[title={ALFWorld}, ymin=0, ymax=115, ytick={0,25,50,75,100}]
\addplot[dropline]    coordinates {(0,100) (0.25,105.7) (0.5,108.4) (0.75,108.7) (1,107.7)};
\addplot[replaceline] coordinates {(0,100) (0.25,105.7) (0.5,104.2) (0.75,99.7) (1,107.5)};
\addplot[paraline]    coordinates {(0,100) (0.25,98.9) (0.5,101.7) (0.75,101.4) (1,102.3)};

% ---------------- Row 2: Action corruption ----------------
\nextgroupplot[
  ylabel={Action\\retention (\%)}, ylabel style={align=center},
  ymin=0, ymax=115, ytick={0,25,50,75,100},
]
\addplot[dropline]    coordinates {(0,100) (0.25,100.4) (0.5,96.7) (0.75,53.8) (1,53.1)};
\addplot[replaceline] coordinates {(0,100) (0.25,102.9) (0.5,97.5) (0.75,81.7) (1,10.4)};

\nextgroupplot[ymin=0, ymax=115, ytick={0,25,50,75,100}]
\addplot[dropline]    coordinates {(0,100) (0.25,97.5) (0.5,95.9) (0.75,92.4) (1,79.5)};
\addplot[replaceline] coordinates {(0,100) (0.25,101.3) (0.5,100.3) (0.75,94.3) (1,46.2)};

\nextgroupplot[ymin=0, ymax=115, ytick={0,25,50,75,100}]
\addplot[dropline]    coordinates {(0,100) (0.25,91.5) (0.5,73.6) (0.75,30.8) (1,0.0)};
\addplot[replaceline] coordinates {(0,100) (0.25,104.5) (0.5,88.1) (0.75,57.2) (1,0.0)};

\nextgroupplot[ymin=0, ymax=115, ytick={0,25,50,75,100}]
\addplot[dropline]    coordinates {(0,100) (0.25,100.5) (0.5,102.5) (0.75,67.4) (1,0.0)};
\addplot[replaceline] coordinates {(0,100) (0.25,100.5) (0.5,102.9) (0.75,65.3) (1,0.0)};
\end{groupplot}
 
% ---- shared legend on top ----
\node[anchor=south] at
  ([yshift=25pt]$(group c1r1.north)!0.5!(group c4r1.north)$)
  {\pgfplotslegendfromname{sweeplegend}};
 
\end{tikzpicture}
\caption{Retention under thought corruption (top) and action corruption (bottom) at varying corruption ratios $p$ (Qwen2.5-3B-Instruct student). Corruption is applied at the span level: each thought $t_i$ or action $a_i$ is dropped, replaced, or paraphrased as a whole with probability $p$. Paraphrase is defined for thoughts only.}
\label{fig:corruption_sweep}
\end{figure}

\paragraph{Thought and action corruption across corruption ratios
(Section~\ref{sec:action-matters}).}
Figure~\ref{fig:corruption_sweep} reports retention under span-level corruption of
thoughts (top row) and actions (bottom row) across corruption ratios
$p \in \{0.25, 0.5, 0.75, 1.0\}$. Thought retention stays near the clean baseline at
every ratio and for all three operators, whereas action retention decays with $p$ and collapses to $0\%$ on the interactive benchmarks at
$p=1.0$. These results complement the full-corruption comparison by including
settings in which only a fraction of spans is corrupted.

\paragraph{Robustness across teacher models (Section~\ref{sec:teacher-model}).} 
Table~\ref{tab:teacher-model} reports the full results under thought drop, replacement, and paraphrase for trajectories generated by Qwen2.5-32B-Instruct, GPT-5.4, Claude-Opus-5, and Gemini-3.1-Pro. The student is fixed to Qwen2.5-3B-Instruct, and each intervention is compared with the clean baseline obtained using the same teacher.

\begin{table}[h]
\centering
\setlength{\tabcolsep}{4pt}
\caption{Task accuracy under thought interventions across teacher models, using a Qwen2.5-3B-Instruct student ($p{=}1.0$).}
\label{tab:teacher-model}
\begin{tabular}{llcccc}
\toprule
Benchmark & Intervention & Qwen2.5-32B-Instruct & GPT-5.4 & Claude-Opus-5 & Gemini-3.1-Pro \\
\midrule
\multirow{4}{*}{HotpotQA}
 & Clean      & 48.0\std{0.7} & 52.6\std{0.9} & 41.0\std{0.6} & 42.8\std{0.9} \\
 & Drop       & 46.8\std{0.3} & 50.0\std{0.6} & 39.2\std{0.4} & 41.6\std{0.6} \\
 & Replace    & 45.6\std{1.5} & 48.8\std{1.5} & 38.4\std{1.4} & 41.6\std{1.1} \\
 & Paraphrase & 46.8\std{1.3} & 52.4\std{1.4} & 41.2\std{0.7} & 46.8\std{0.9} \\
\midrule
\multirow{4}{*}{FEVER}
 & Clean      & 63.4\std{1.2} & 61.0\std{1.0} & 59.2\std{0.8} & 59.0\std{1.2} \\
 & Drop       & 64.9\std{0.8} & 57.4\std{0.9} & 57.2\std{0.7} & 57.4\std{0.9} \\
 & Replace    & 63.9\std{2.5} & 57.0\std{2.2} & 56.4\std{1.7} & 55.2\std{1.9} \\
 & Paraphrase & 62.0\std{1.2} & 58.8\std{1.4} & 58.8\std{1.0} & 59.2\std{1.1} \\
\midrule
\multirow{4}{*}{WebShop}
 & Clean      & 40.2\std{0.9} & 47.8\std{1.0} & 48.6\std{1.0} & 44.6\std{1.0} \\
 & Drop       & 43.1\std{0.8} & 48.2\std{0.9} & 52.2\std{1.0} & 47.0\std{0.8} \\
 & Replace    & 43.7\std{1.0} & 46.8\std{1.3} & 48.6\std{1.2} & 44.8\std{1.3} \\
 & Paraphrase & 37.8\std{0.9} & 45.8\std{1.3} & 48.2\std{0.9} & 46.2\std{1.2} \\
\midrule
\multirow{4}{*}{ALFWorld}
 & Clean      & 88.3\std{1.0} & 96.4\std{0.4} & 98.9\std{0.4} & 99.3\std{0.4} \\
 & Drop       & 95.1\std{0.9} & 98.2\std{0.4} & 99.3\std{0.4} & 97.6\std{1.2} \\
 & Replace    & 94.9\std{1.0} & 98.7\std{0.2} & 99.5\std{0.2} & 99.3\std{0.4} \\
 & Paraphrase & 90.3\std{1.7} & 97.8\std{0.4} & 99.3\std{0.4} & 99.3\std{0.4} \\
\bottomrule
\end{tabular}
\end{table}

\paragraph{Robustness across student model families (Section~\ref{sec:student-family}).} 
Table~\ref{tab:student-family} reports the full results under thought interventions for Llama-3.2-3B-Instruct, Qwen2.5-3B-Instruct, and Phi-4-mini-instruct. All students are distilled from the same Qwen2.5-32B-Instruct teacher trajectories, with a separate clean baseline for each student model.

\begin{table}[h]
\centering
\setlength{\tabcolsep}{3.7pt}
\caption{Task accuracy under thought interventions across student model families, using a fixed Qwen2.5-32B-Instruct teacher ($p{=}1.0$).}
\label{tab:student-family}
\begin{tabular}{llccc}
\toprule
Benchmark & Intervention & Llama-3.2-3B-Instruct & Qwen2.5-3B-Instruct & Phi-4-mini-instruct \\
\midrule
\multirow{4}{*}{HotpotQA}
 & Clean      & 51.2\std{0.7} & 48.0\std{0.7} & 45.4\std{1.3} \\
 & Drop       & 49.4\std{0.7} & 46.8\std{0.3} & 42.4\std{1.0} \\
 & Replace    & 47.6\std{1.6} & 45.6\std{1.5} & 42.4\std{2.0} \\
 & Paraphrase & 51.6\std{1.0} & 46.8\std{1.3} & 47.2\std{1.3} \\
\midrule
\multirow{4}{*}{FEVER}
 & Clean      & 67.6\std{0.8} & 63.4\std{1.2} & 64.8\std{1.0} \\
 & Drop       & 66.0\std{1.1} & 64.9\std{0.8} & 65.4\std{1.5} \\
 & Replace    & 66.6\std{2.3} & 63.9\std{2.5} & 62.2\std{3.1} \\
 & Paraphrase & 64.2\std{1.2} & 62.0\std{1.2} & 62.4\std{1.7} \\
\midrule
\multirow{4}{*}{WebShop}
 & Clean      & 44.6\std{0.8} & 40.2\std{0.9} & 42.2\std{1.0} \\
 & Drop       & 47.0\std{1.0} & 43.1\std{0.8} & 42.5\std{1.3} \\
 & Replace    & 46.4\std{1.3} & 43.7\std{1.0} & 39.6\std{1.7} \\
 & Paraphrase & 42.2\std{1.1} & 37.8\std{0.9} & 39.8\std{1.4} \\
\midrule
\multirow{4}{*}{ALFWorld}
 & Clean      & 94.2\std{0.7} & 88.3\std{1.0} & 91.6\std{1.0} \\
 & Drop       & 98.2\std{1.1} & 95.1\std{0.9} & 94.9\std{1.8} \\
 & Replace    & 97.8\std{1.3} & 94.9\std{1.0} & 95.6\std{2.2} \\
 & Paraphrase & 92.7\std{1.1} & 90.3\std{1.7} & 88.0\std{1.6} \\
\bottomrule
\end{tabular}
\end{table}

\paragraph{Robustness to corruption granularity
(Section~\ref{sec:token-level}).}
Table~\ref{tab:span-vs-token} reports the full corruption-ratio sweeps for thought drop, replacement, and paraphrase at both span and token levels. At the token level, each thought token is independently selected with probability $p$. For replacement, selected tokens are independently sampled from a deduplicated token pool constructed from thoughts in other trajectories. For paraphrase, we instead prompt the Qwen2.5-32B-Instruct teacher with the surrounding thought context to generate a context-appropriate synonym for each selected token, preserving the local meaning while altering its lexical realization. For drop, we remove the selected tokens.

\begin{table}[h]
\centering
\setlength{\tabcolsep}{5pt}
\caption{Task accuracy under span-level and token-level thought corruption at varying corruption ratios $p$ (Qwen2.5-3B-Instruct student, Qwen2.5-32B-Instruct teacher). $p=0$ corresponds to the unperturbed trajectories.}
\label{tab:span-vs-token}
\begin{tabular}{llcccccc}
\toprule
& & \multicolumn{2}{c}{Drop} & \multicolumn{2}{c}{Replace} & \multicolumn{2}{c}{Paraphrase} \\
\cmidrule(lr){3-4}\cmidrule(lr){5-6}\cmidrule(lr){7-8}
Benchmark & $p$ & Span & Token & Span & Token & Span & Token \\
\midrule
\multirow{5}{*}{HotpotQA}
 & 0 (clean) & \multicolumn{6}{c}{48.0} \\
 & 0.25 & 48.2 & 47.8 & 47.6 & 48.4 & 48.2 & 47.0 \\
 & 0.50 & 47.6 & 47.6 & 47.2 & 46.0 & 49.0 & 47.4 \\
 & 0.75 & 45.6 & 44.7 & 45.4 & 45.0 & 49.8 & 50.6 \\
 & 1.00 & 46.8 & 46.4 & 45.6 & 44.8 & 46.8 & 48.0 \\
\midrule
\multirow{5}{*}{FEVER}
 & 0 (clean) & \multicolumn{6}{c}{63.4} \\
 & 0.25 & 65.8 & 62.4 & 61.6 & 64.0 & 64.4 & 61.8 \\
 & 0.50 & 63.6 & 62.2 & 64.2 & 63.2 & 62.2 & 64.0 \\
 & 0.75 & 63.6 & 62.5 & 60.0 & 62.2 & 62.2 & 64.8 \\
 & 1.00 & 64.9 & 64.3 & 63.9 & 61.8 & 62.0 & 65.2 \\
\midrule
\multirow{5}{*}{WebShop}
 & 0 (clean) & \multicolumn{6}{c}{40.2} \\
 & 0.25 & 41.6 & 41.4 & 42.4 & 40.4 & 41.6 & 39.0 \\
 & 0.50 & 44.2 & 40.6 & 42.8 & 40.2 & 39.8 & 39.0 \\
 & 0.75 & 40.0 & 43.2 & 44.0 & 45.0 & 41.0 & 39.8 \\
 & 1.00 & 43.1 & 42.6 & 43.7 & 45.2 & 37.8 & 40.0 \\
\midrule
\multirow{5}{*}{ALFWorld}
 & 0 (clean) & \multicolumn{6}{c}{88.3} \\
 & 0.25 & 93.3 & 92.3 & 93.3 & 91.6 & 87.3 & 90.1 \\
 & 0.50 & 95.7 & 93.4 & 92.0 & 92.7 & 89.8 & 93.8 \\
 & 0.75 & 96.0 & 93.4 & 88.0 & 92.3 & 89.5 & 93.4 \\
 & 1.00 & 95.1 & 94.0 & 94.9 & 95.3 & 90.3 & 89.8 \\
\bottomrule
\end{tabular}
\end{table}

\paragraph{Robustness across distillation objectives
(Section~\ref{sec:exp-obj}).}
Table~\ref{tab:kl} compares SFT and KL-based distillation under span-level thought drop at matched corruption ratios. For each objective, the result at $p=0$ serves as the clean baseline for assessing the effect of thought removal.

\begin{table}[ht]
\centering
\caption{Accuracy under SFT and KL distillation objectives as a function of the span-level thought-drop ratio $p$ (Qwen2.5-3B-Instruct student, Qwen2.5-32B-Instruct teacher). $p{=}0$ corresponds to the unperturbed trajectories.}
\label{tab:kl}
\begin{tabular}{llcccc}
\toprule
Objective & $p$ & HotpotQA & FEVER & WebShop & ALFWorld \\
\midrule
\multirow{5}{*}{SFT} & 0 (clean) & 48.0 & 63.4 & 40.2 & 88.3 \\
 & 0.25 & 48.2 & 65.8 & 41.6 & 93.3 \\
 & 0.50 & 47.6 & 63.6 & 44.2 & 95.7 \\
 & 0.75 & 45.6 & 63.6 & 40.0 & 96.0 \\
 & 1.00 & 46.8 & 64.9 & 43.1 & 95.1 \\
\midrule
\multirow{5}{*}{KL}  & 0 (clean) & 43.8 & 61.6 & 39.0 & 89.1 \\
                     & 0.25      & 43.4 & 61.8 & 39.0 & 87.1 \\
                     & 0.50      & 43.0 & 60.6 & 38.2 & 86.9 \\
                     & 0.75      & 41.6 & 60.0 & 37.8 & 89.1 \\
                     & 1.00      & 40.8 & 59.2 & 37.8 & 87.3 \\
\bottomrule
\end{tabular}
\end{table}

\paragraph{Robustness to out-of-distribution datasets
(Section~\ref{sec:ood}).}
Table~\ref{tab:ood-robustness} reports the full results for a Qwen2.5-14B-Instruct student distilled on HotpotQA under thought drop, replacement, and paraphrase at $p=1.0$. Evaluation includes HotpotQA as the in-distribution benchmark and FEVER, 2WikiMultiHopQA, Bamboogle, and MuSiQue as out-of-distribution benchmarks, without target-dataset fine-tuning. Each intervention is compared with the clean model on the same evaluation dataset.

\begin{table}[ht]
\centering
\caption{In-distribution and out-of-distribution performance of a Qwen2.5-14B-Instruct student distilled on HotpotQA under thought drop, replacement, and paraphrase ($p{=}1.0$).}
\label{tab:ood-robustness}
% \small
\setlength{\tabcolsep}{6pt}
\begin{tabular}{lcccc}
\toprule
Evaluation Dataset
& Clean
& Thought Drop
& Thought Replace
& Thought Paraphrase \\
\midrule
HotpotQA (In-distribution)
& 56.4
& 58.6
& 52.4 
& 55.0 \\

FEVER (Out-of-distribution)
& 63.2
& 65.2
& 65.8
& 62.8 \\

2WikiMultiHopQA (Out-of-distribution)
& 49.4
& 50.0 
& 47.4 
& 46.8 \\

Bamboogle (Out-of-distribution)
& 50.4
& 52.0
& 47.2
& 50.4 \\

MuSiQue (Out-of-distribution)
& 21.2
& 21.6
& 19.8
& 23.8 \\
\bottomrule
\end{tabular}
\end{table}

\paragraph{Thought interventions in low-disclosure settings (Section~\ref{sec:low-disclosure}).} 
Table~\ref{tab:finqa-thought-drop} reports accuracy under clean training, thought drop, and thought replacement on FinQA, MultiHiertt, MathQA, and GeoQA across four student scales. Both interventions reduce accuracy in every reported benchmark--scale combination, with larger students also remaining sensitive to the removal or replacement of thoughts.

\begin{table}[h]
\centering
\caption{Accuracy under clean training, thought drop, and thought replacement across Qwen2.5-\{0.5B, 3B, 7B, 14B\}-Instruct students on low-disclosure settings, using a Qwen2.5-32B-Instruct teacher ($p{=}1.0$).}
\label{tab:finqa-thought-drop}
\begin{tabular}{llccc}
\toprule
& & & \multicolumn{2}{c}{Thought} \\
\cmidrule(lr){4-5}
Benchmark & Size & Clean & Drop & Replace \\
\midrule
\multirow{4}{*}{FinQA}
 & 0.5B & 52.7\std{0.6} & \phantom{0}3.0\std{0.3} & \phantom{0}3.0\std{0.3} \\
 & 3B   & 63.3\std{0.7} & 47.7\std{1.5} & 49.0\std{1.9} \\
 & 7B   & 68.0\std{0.6} & 54.4\std{1.2} & 52.9\std{1.7} \\
 & 14B  & 72.7\std{0.7} & 56.2\std{1.6} & 56.6\std{1.3} \\
\midrule
\multirow{4}{*}{MultiHiertt}
 & 0.5B & 32.3\std{1.2} & 23.8\std{1.3} & 21.6\std{1.8} \\
 & 3B   & 49.2\std{0.6} & 39.5\std{1.1} & 40.5\std{1.6} \\
 & 7B   & 55.3\std{0.8} & 46.0\std{0.9} & 46.4\std{1.4} \\
 & 14B  & 60.1\std{0.6} & 50.4\std{1.2} & 50.1\std{1.1} \\
\midrule
\multirow{4}{*}{MathQA}
 & 0.5B & \phantom{0}6.0\std{0.3} & \phantom{0}1.1\std{0.2} & \phantom{0}3.0\std{0.7} \\
 & 3B   & 17.1\std{0.9} & 10.7\std{0.8} & 10.3\std{1.2} \\
 & 7B   & 29.5\std{0.6} & 13.2\std{1.0} & 17.2\std{1.5} \\
 & 14B  & 41.2\std{1.2} & 13.5\std{1.0} & 17.8\std{1.9} \\
\midrule
\multirow{4}{*}{GeoQA}
 & 0.5B & 25.2\std{0.4} & 5.8\std{0.6} & 4.0\std{0.3} \\
 & 3B   & 38.8\std{0.6} & 20.1\std{0.5} & 21.9\std{0.7} \\
 & 7B   & 52.3\std{0.5} & 22.3\std{0.7} & 30.2\std{0.6} \\
 & 14B  & 56.7\std{0.4} & 30.6\std{0.6} & 37.0\std{0.5} \\
\bottomrule
\end{tabular}
\end{table}

\paragraph{Training and inference efficiency (Section~\ref{sec:solution}).} 
Table~\ref{tab:efficiency_seconds} reports the absolute wall-clock training and inference times for full-trajectory and action-only distillation across the high-disclosure settings. Time reductions are computed relative to full-trajectory distillation for each benchmark.

\begin{table}[ht]
\centering
\setlength{\tabcolsep}{4pt}
\caption{Wall-clock training and inference times under full-trajectory and action-only distillation. Absolute times are reported in seconds, and reductions are relative to full-trajectory distillation.}
\label{tab:efficiency_seconds}
\begin{tabular}{lrrrrrr}
\toprule
& \multicolumn{3}{c}{Training}
& \multicolumn{3}{c}{Inference} \\
\cmidrule(lr){2-4}
\cmidrule(lr){5-7}
Benchmark
& \multicolumn{1}{c}{Full (s)} & \multicolumn{1}{c}{Action-only (s)} & \multicolumn{1}{c}{Reduction (\%)}
& \multicolumn{1}{c}{Full (s)} & \multicolumn{1}{c}{Action-only (s)} & \multicolumn{1}{c}{Reduction (\%)} \\
\midrule
HotpotQA & $441.3 $\std{11.8} & $410.0 $\std{1.0} & 7.1  & $687.8 $\std{5.2} & $492.6 $\std{1.2} & 28.4 \\
FEVER    & $377.3 $\std{4.2} & $353.7 $\std{2.3} & 6.3  & $361.7 $\std{1.5} & $321.7 $\std{2.1} & 11.1 \\
ALFWorld & $2744.0 $\std{21.3}& $2463.7 $\std{21.4}& 10.2 & $891.0 $\std{34.7}& $519.3 $\std{5.7} & 41.7 \\
WebShop  & $1890.3 $\std{15.9}& $1791.0 $\std{8.5}& 5.3  & $608.3 $\std{13.0}& $512.0 $\std{19.0}& 15.8 \\
\bottomrule
\end{tabular}
\end{table}

\section{Details of the Disclosure Intervention}
\label{app:disclosure}

\subsection{Thought Generation for Low-disclosure Trajectories}
\label{app:low-disclosure-thought-generation}
For the low-disclosure settings, the accompanying thoughts are generated with Qwen2.5-32B-Instruct after constructing and verifying the corresponding executable action. The model is provided with the task input and the verified step-wise computation underlying the action, including its intermediate values and final result. As shown in Figure~\ref{fig:low-disclosure-thought-prompt}, it is instructed to narrate this computation concisely in the first person, as if working through the problem for the first time, without solving it anew.

\begin{figure}[ht]
\begin{lstlisting}[style=promptbox]
=== System prompt ===
You write short first-person reasoning narration for a math problem-solving agent. You are given the problem 
and the EXACT correct arithmetic already worked out. Do not solve anything -- just narrate it naturally and 
concisely, as if you were figuring it out for the first time.

=== User prompt ===
Problem: {question}

The correct calculation (already verified correct) is:
{trace}

Final result: {final_result}

Write short reasoning snippets as this agent would say them.
\end{lstlisting}
\caption{Prompt used to generate thoughts for low-disclosure trajectories.}
\label{fig:low-disclosure-thought-prompt}
\end{figure}

\subsection{Semantic Naming Procedure}
\label{app:semantic-naming}

For the disclosure intervention in Section~\ref{sec:disclose}, we transform the executable actions of the low-disclosure settings while keeping the task input, thought, numerical computation, and evaluation instances fixed. Starting from the original action containing anonymous intermediate variables (e.g., \texttt{step0}, \texttt{step1}), we first execute the original code to obtain its output. We then prompt Qwen2.5-32B-Instruct to replace each anonymous variable with a descriptive, question-grounded identifier. The model is explicitly instructed not to change numerical constants, operators, or the order of operations. We decode with temperature $0.2$ and a maximum of $320$ tokens.

To verify the transformation, we re-execute each transformed action and confirm that its output matches that of the original program within a relative tolerance of $10^{-6}$. We additionally require the rewritten code to include at least one assigned variable whose identifier no longer matches the anonymous \texttt{step\textbackslash d+} pattern. For accepted examples, only the code block in the original trajectory is replaced; the question, context, and thought remain unchanged. The same evaluation instances are used for the original and disclosed conditions.

\begin{figure}[ht]
\begin{lstlisting}[style=promptbox]
=== System prompt ===
You rewrite Python snippets to be readable. You are given a quantitative reasoning problem and a correct 
snippet whose variables are called step0, step1, ... Rename every variable to a descriptive name drawn from the 
question. NEVER change any number, any operator, or the order of operations. Output only the rewritten code, no 
fences and no explanation.

=== User prompt ===
Question: {question}

Snippet:
{code}

Rewritten:
\end{lstlisting}
\caption{Prompt used for semantic naming in the disclosure intervention.}
\label{fig:semantic-naming-prompt}
\end{figure}

\subsection{Mismatched-Name Control}
\label{app:disclosure_control}

The intervention in Section~\ref{sec:disclose} introduces semantically informative identifiers into the action representation. To further examine the role of semantic correspondence, we construct a \emph{mismatched-name} control in which each anonymous variable is renamed using content words from the corresponding thought that do not describe the quantity stored in that variable (e.g., naming $15 - 10$ as \texttt{percentage\_of\_meters} in Figure~\ref{fig:disclosure-intervention-main}). This preserves thought-derived variable names while breaking their correspondence with the represented quantities. Numeric values, operators, execution order, and final outputs remain unchanged, so the underlying computation is identical across representations.

Table~\ref{tab:mismatched_control} shows that mismatched variable names do not reproduce the benefit of semantically informative identifiers. Across all four benchmarks, thought-drop retention under the mismatched condition remains below that of the semantic condition. Clean accuracy is consistently lower than in the anonymous condition, suggesting that misleading identifiers may interfere with learning from the demonstrated actions. Together, these results are consistent with semantic alignment between identifiers and their represented quantities contributing to the improved retention under the semantic condition.

\begin{table}[h]
\centering
\caption{Thought-drop retention under the anonymous, semantic, and mismatched-name action representations in the low-disclosure settings, using a Qwen2.5-3B-Instruct student. $\Delta$ denotes the absolute accuracy change from Clean to Drop (percentage points).}
\label{tab:mismatched_control}
\begin{tabular}{llcccc}
\toprule
Benchmark & Action representation & Clean & Drop & $\Delta$ & Retention (\%) \\
\midrule
\multirow{3}{*}{FinQA}
 & Anonymous (= before disclosing) & 63.3 & 47.7 & $-$15.6 & 75.4 \\
 & Semantic (= after disclosing)   & 65.3 & 61.8 & $-$3.5  & 94.6 \\
 & Mismatched                      & 45.3 & 36.5 & $-$8.8  & 80.6 \\
\midrule
\multirow{3}{*}{MultiHiertt}
 & Anonymous (= before disclosing) & 49.2 & 39.5 & $-$9.7  & 80.3 \\
 & Semantic (= after disclosing)   & 50.6 & 44.6 & $-$6.0  & 88.1 \\
 & Mismatched                      & 33.8 & 26.3 & $-$7.5  & 77.8 \\
\midrule
\multirow{3}{*}{MathQA}
 & Anonymous (= before disclosing) & 17.1 & 10.7 & $-$6.4  & 62.6 \\
 & Semantic (= after disclosing)   & 18.3 & 16.7 & $-$1.6  & 91.3 \\
 & Mismatched                      & 16.7 & 12.7 & $-$4.0  & 76.0 \\
\midrule
\multirow{3}{*}{GeoQA}
 & Anonymous (= before disclosing) & 38.8 & 20.1 & $-$18.7 & 51.8 \\
 & Semantic (= after disclosing)   & 41.0 & 34.6 & $-$6.4  & 84.4 \\
 & Mismatched                      & 26.8 & 16.3 & $-$10.5 & 60.8 \\
\bottomrule
\end{tabular}
\end{table}

\section{Controlling for Inference-Time Thought Generation}
\label{app:inference_control}

Our thought-drop intervention changes not only the supervision available during training but also the student's generation behavior at inference time: full-trajectory students generate a thought before each action, whereas students trained without thoughts produce actions directly. The performance gap on low-disclosure settings (Section~\ref{sec:low-disclosure}) therefore admits two explanations: teacher thoughts may provide training signal that cannot be learned from actions alone, or full-trajectory students may simply benefit from additional natural-language reasoning at inference time. To separate these two factors, we examine whether full-trajectory students retain their advantage when thought generation is suppressed at inference time.

Specifically, we take the full-trajectory students trained with the original anonymous action representation and evaluate them under a direct-action inference setting (\textsc{Full + Direct}). The model parameters are kept unchanged; only the inference prompt is modified to request the action directly without an intermediate thought. Since neither \textsc{Full + Direct} nor \textsc{Action-only} generates thoughts at inference time, their performance difference cannot be explained by additional inference-time thought generation.

As shown in Table~\ref{tab:inference_control}, \textsc{Full + Direct} lies between \textsc{Action-only} and \textsc{Full + Thought} on all four benchmarks, recovering $41.2$--$67.2\%$ of the gap between them. This estimate may be conservative, as \textsc{Full + Direct} is evaluated in an inference format unseen during training. These results show that thought supervision provides benefits beyond inference-time thought generation: full-trajectory students consistently outperform action-only students even when both predict actions directly.

\begin{table}[t]
\centering
\caption{Accuracy of full-trajectory-trained students, with and without
inference-time thought generation, and action-only students in the low-disclosure settings. \textsc{Full + Thought} uses the standard thought--action inference format,
whereas \textsc{Full + Direct} suppresses thought generation through the prompt
and directly predicts the action. \textsc{Action-only} denotes students trained without thoughts. All results use Qwen2.5-3B-Instruct students with the
anonymous action representation.}
\label{tab:inference_control}
\begin{tabular}{lccc}
\toprule
Benchmark & \textsc{Full + Thought} & \textsc{Full + Direct} & \textsc{Action-only} \\
\midrule
FinQA       & 63.3\std{0.7} & 57.1\std{1.0} & 47.7\std{1.5} \\
MultiHiertt & 49.2\std{0.6} & 43.5\std{0.9} & 39.5\std{1.1} \\
MathQA      & 17.1\std{0.9}& 15.0\std{1.2} & 10.7\std{0.8} \\
GeoQA       & 38.8\std{0.6} & 32.0\std{0.7} & 20.1\std{0.5} \\
\bottomrule
\end{tabular}
\end{table}

\section{Robustness in a Low-Level Action Environment}
\label{sec:low_level_actions}

To further examine whether the robustness to thought removal generalizes to environments with more fine-grained action spaces, we additionally evaluate our approach on InterCode~\citep{yang2023intercode}, where agents interact with the environment through low-level executable actions. Unlike the high-level natural-language actions used in benchmarks such as WebShop and ALFWorld, InterCode requires agents to produce more fine-grained action sequences. We therefore use InterCode to test whether the limited dependence on thought supervision observed in our main experiments also extends to low-level action environments.

We apply the same distillation setup and compare students trained on the complete teacher trajectories with those trained after removing all teacher thoughts. As shown in Table~\ref{tab:intercode}, thought removal preserves most of the clean performance across all four student scales. Retention is 104.8\%, 101.2\%, 97.7\%, and 95.3\% for the 0.5B, 3B, 7B, and 14B students, respectively, with an average retention of 99.8\%. The 0.5B and 3B students slightly exceed their clean baselines, while the 7B and 14B students show only modest degradation. 

These results show that the robustness to thought removal is not restricted to benchmarks with high-level actions, but also extends to an environment requiring substantially more fine-grained, low-level actions. This further supports our finding that explicit thought supervision is not always necessary for transferring effective agent behavior.

\begin{table}[h]
\centering
\caption{Robustness to thought removal on InterCode, a low-level action environment.}
\label{tab:intercode}
\begin{tabular}{lccc}
\toprule
Student & Clean & Thought Drop & Retention (\%) \\
\midrule
Qwen2.5-0.5B-Instruct & 0.525 & 0.550 & 104.8 \\
Qwen2.5-3B-Instruct   & 0.588 & 0.595 & 101.2 \\
Qwen2.5-7B-Instruct   & 0.563 & 0.550 & 97.7 \\
Qwen2.5-14B-Instruct  & 0.485 & 0.462 & 95.3 \\
\bottomrule
\end{tabular}
\end{table}

\end{document}